\documentclass[acmsmall,nonacm]{acmart}
\usepackage{graphicx}
\usepackage{amsthm}
\usepackage{amsmath}

\AtBeginDocument{%
  }

\setcopyright{acmlicensed}
\copyrightyear{2025}
\acmYear{2025}
\acmJournal{JACM}

\begin{document}

\title{AI Agent Systems: Architectures, Applications, and Evaluation}

% (Optional) Add authors back later if needed:
% \author{...}
\author{Bin Xu}
\orcid{0009-0001-2639-7283}
\affiliation{%
  \institution{School of Electrical, Computer and Energy Engineering, Arizona State University}
  \city{Tempe}
  \state{Arizona}
  \country{USA}
}
\email{binxu4@asu.edu}

\begin{abstract}
AI agents---systems that combine foundation models with reasoning, planning, memory, and tool use---are rapidly becoming a practical interface between natural-language intent and real-world computation. This survey synthesizes the emerging landscape of AI agent architectures for (i) deliberation and reasoning (e.g., chain-of-thought style decomposition, self-reflection and verification, and constraint-aware decision making), (ii) planning and control (from reactive policies to hierarchical and multi-step planners), and (iii) tool calling and environment interaction (retrieval, code execution, APIs, and multimodal perception). We organize prior work into a unified taxonomy spanning agent components (policy/LLM core, memory, world models, planners, tool routers, and critics), orchestration patterns (single-agent vs.\ multi-agent; centralized vs.\ decentralized coordination), and deployment settings (offline analysis vs.\ online interactive assistance; safety-critical vs.\ open-ended tasks). We discuss key design trade-offs---latency vs.\ accuracy, autonomy vs.\ controllability, and capability vs.\ reliability---and highlight how evaluation is complicated by non-determinism, long-horizon credit assignment, tool and environment variability, and hidden costs such as retries and context growth. Finally, we summarize measurement and benchmarking practices (task suites, human preference and utility metrics, success under constraints, robustness and security) and identify open challenges including verification and guardrails for tool actions, scalable memory and context management, interpretability of agent decisions, and reproducible evaluation under realistic workloads.
\end{abstract}


\maketitle

\section{Introduction}
\subsection{Motivation}
Foundation models have made natural language a practical interface for computation, but most real tasks are not single-turn question answering. They involve gathering information from multiple sources, maintaining state over time, choosing among tools, and executing multi-step actions under constraints (latency, permissions, safety, and cost). \emph{AI agents} address this gap by coupling a foundation model with an execution loop that can observe an environment, plan, call tools, update memory, and verify outcomes \cite{masterman2024agentarchitectures,chowa2025languagetoaction}. In other words, an agent is not only a generator of text; it is a controller that translates intent into \emph{procedures} carried out in the world (software repositories, browsers, enterprise systems, or physical robots).

\subsection{Background}
Modern digital work is fragmented across interfaces and APIs: knowledge is distributed (documents, databases, dashboards), actions are mediated by tools (search, code execution, ticketing systems), and success is defined by end-to-end outcomes rather than plausibility. Purely conversational systems often fail in these settings due to hallucinations, lack of grounding, and inability to execute or verify actions. Tool-augmented and retrieval-augmented designs improve reliability by binding claims to evidence and by making intermediate artifacts inspectable \cite{lewis2020rag,yao2023react}. Modular tool routing (e.g., MRKL-style) further improves governance by separating language understanding from specialized tools and by enforcing structured interfaces that can be audited \cite{karpas2022mrkl,schick2023toolformer}.

\subsection{Overview}
Agents are especially important in the current age for three reasons. First, the scope of tasks is expanding from writing assistance to \emph{workflow automation}: coding agents resolve issues end-to-end \cite{jimenez2023swebench,sweagent2024}, web agents operate real sites under variability \cite{zhou2023webarena,mind2web2023,webshop2022}, and enterprise assistants orchestrate multi-step operations with policy constraints. Second, deployment is increasingly \emph{interactive and long-horizon}, where small errors compound and nondeterminism (sampling, tool failures) complicates reproducibility, motivating verification loops and trace-based evaluation \cite{qin2023toolbench,liu2023agentbench,yehudai2025agentseval}. Third, safety and security pressures are rising: prompt injection, untrusted retrieved content, and side-effecting tools require defense-in-depth alignment and guardrails beyond the final response \cite{bai2022constitutional}.

\begin{figure}[thbp]
\centering
\includegraphics[width=0.8\linewidth]{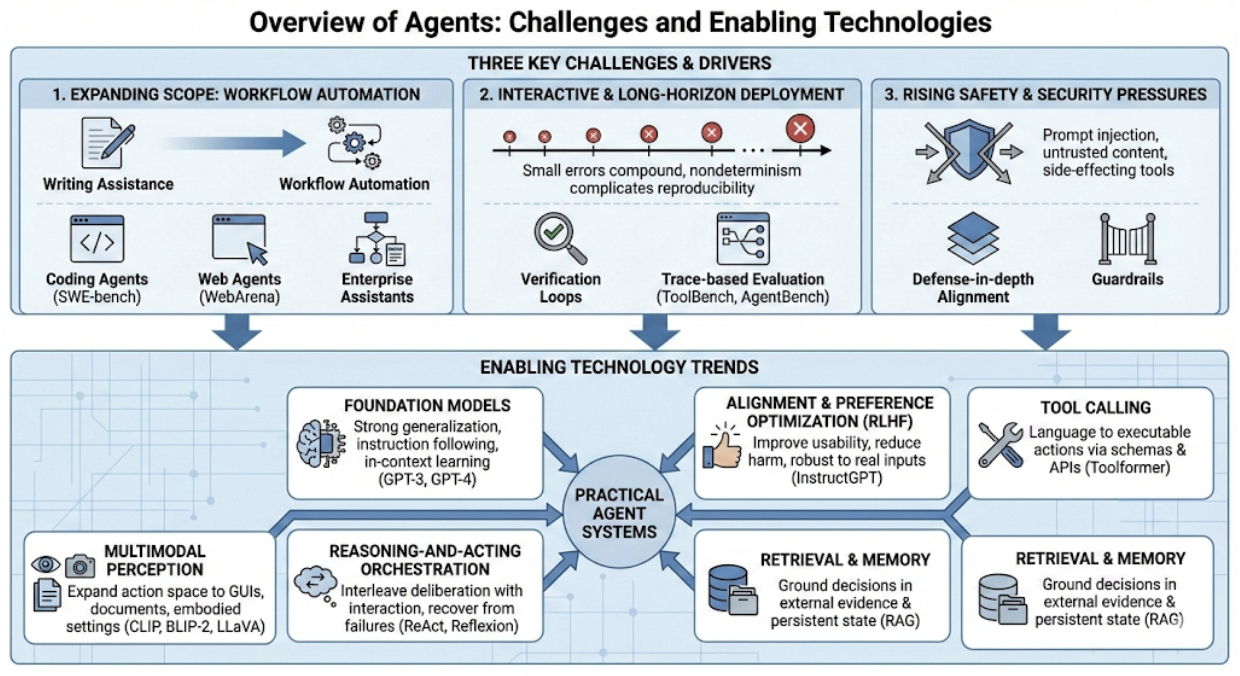}
\caption{Overview of AI agents and the agent execution loop (reasoning, tools, and memory)}
\label{fig:section1}
\end{figure}

Fig.~\ref{fig:section1} provides a high-level visual overview of the main components and execution loop of an AI agent.

Several technology trends enable practical agent systems today. \textbf{Foundation models} provide strong generalization, instruction following, and emergent in-context learning that supports rapid adaptation without retraining \cite{brown2020gpt3,openai2023gpt4}. \textbf{Alignment and preference optimization} (e.g., RLHF) improve usability and reduce harmful behavior, making agents more robust under real user inputs \cite{christiano2017rlhf,ouyang2022instructgpt}. \textbf{Tool calling} turns language into executable actions via schemas and APIs \cite{schick2023toolformer,patil2023gorilla}, while \textbf{retrieval and memory} ground decisions in external evidence and persistent state \cite{schick2023toolformer,lewis2020rag,memgpt2023}. \textbf{Reasoning-and-acting orchestration} interleaves deliberation with environment interaction to improve grounding and recover from failures \cite{yao2023react,shinn2023reflexion}. Finally, \textbf{multimodal perception} expands the action space to GUIs, documents, and embodied settings by grounding language in visual inputs \cite{radford2021clip,li2023blip2,liu2023llava}.

\subsection{Current Gaps}
Despite rapid progress, agent systems remain limited by reliability, reproducibility, and governance at scale. Long-horizon tasks amplify compounding errors, and nondeterminism (sampling, tool variability) makes evaluation and debugging difficult without standardized protocols and trace completeness \cite{qin2023toolbench,liu2023agentbench,luo2025llmagentsurvey}. Tool-centric agents also introduce new safety and security risks: untrusted retrieved content and prompt injection can manipulate tool use, and side-effecting actions require stronger constraints than text-only moderation \cite{bai2022constitutional,karpas2022mrkl,sang2025beyondpipelines}. Finally, system-level trade-offs---autonomy vs.\ controllability, latency vs.\ reliability, and capability vs.\ safety---are not yet well understood across domains and deployment settings \cite{sapkota2025agentsvsagentic,zhou2024taxonomyarchexoptions}.

This survey synthesizes emerging agent architectures for reasoning, planning, tool use, and deployment. We organize the landscape along (i) \emph{learning strategies} and system optimization (\S\ref{sec:learning}), and (ii) \emph{application tasks} that stress different capabilities and evaluation regimes (\S\ref{sec:applications}). Throughout, we highlight recurring design trade-offs and emphasize reproducible evaluation under realistic tool and environment variability.

\begin{figure}[thbp]
\centering
\includegraphics[width=0.8\linewidth]{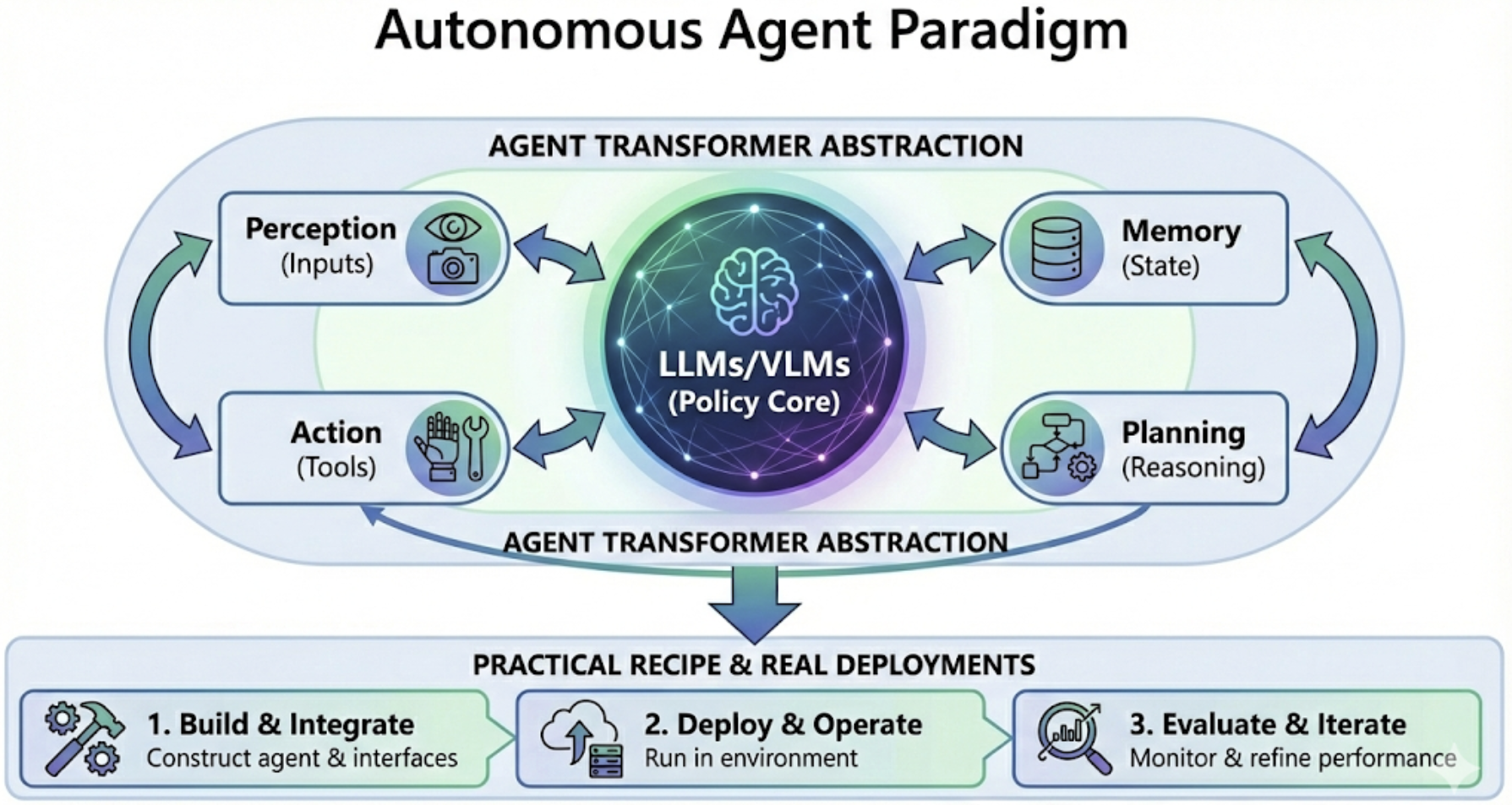}
\caption{Agent-centric AI paradigm: models embedded in tool- and environment-interaction loops}
\label{fig:section2}
\end{figure}

Fig.~\ref{fig:section2} summarizes the agent-centric paradigm that motivates the architectural and evaluation choices discussed in the remainder of the survey.

\section{Autonomous Agent Paradigm}
\label{sec:paradigm}
This section introduces a unifying paradigm for agent systems centered on transformer-based foundation models. We first summarize the role of LLMs/VLMs as the policy core, then define an ``agent transformer'' abstraction that makes agent components and interfaces explicit, and finally describe a practical recipe for building agent transformers in real deployments \cite{microsoft2024agentai_position,sang2025beyondpipelines,petrova2025webofagents}.

\subsection{LLMs and VLMs}
Large Language Models (LLMs) are the dominant \emph{policy cores} for modern agents: they map heterogeneous context (instructions, retrieved documents, tool outputs, and internal memory) to decisions (plans, tool calls, or natural-language actions). Frontier-scale models exhibit strong instruction following and in-context learning, enabling rapid capability bootstrapping without retraining \cite{brown2020gpt3,openai2023gpt4}. However, LLMs are not inherently grounded: without external evidence and executable checks they can hallucinate plausible but incorrect statements. This motivates tool-centric and retrieval-centric agent designs where the model is an orchestrator over trusted tools and data sources \cite{lewis2020rag,yao2023react,karpas2022mrkl}.

An important recent shift is that capability gains increasingly come from \emph{system design} rather than only from bigger backbones. Modern deployments treat the LLM as a planner/controller inside a budgeted loop: the agent is constrained by explicit limits on time, tokens, tool calls, and permissible side effects, and it dynamically allocates ``thinking'' (deliberation) only when the task is hard or risky. This connects directly to test-time compute scaling: self-consistency, reranking, backtracking, and tree-style search can improve reliability without retraining, but must be used selectively to avoid runaway cost and latency \cite{wang2022selfconsistency,yao2023tree}.
Relatedly, agents increasingly rely on \emph{structured action spaces} (typed tool schemas and structured outputs) as the primary control surface: the model proposes actions that must pass schema validation and policy checks before execution, reducing the impact of free-form hallucinations and enabling stronger auditing \cite{schick2023toolformer,karpas2022mrkl}.
Finally, the practical frontier is shifting from ``answering'' to ``operating'': agents are expected to maintain state, recover from tool failures, and justify actions with evidence traces, which places greater emphasis on memory design and trace completeness as first-class artifacts \cite{yao2023react,liu2023agentbench,qin2023toolbench}.

Vision-Language Models (VLMs) extend this paradigm by grounding decisions in images, screens, documents, and embodied observations. Contrastive and instruction-tuned VLMs provide a robust interface from pixels to tokens, enabling agents to operate GUIs (screenshots), read charts and forms, and align actions with visual state \cite{radford2021clip,li2023blip2,liu2023llava}. In practice, multimodal agents often decompose perception into tools (OCR, detection, layout parsing) and use an LLM as the planner/controller that integrates visual evidence with text and tool outputs \cite{yao2023react,openai2023gpt4}. This division improves auditability: intermediate perceptual artifacts can be inspected and verified before committing to downstream actions.

Alignment and preference optimization are also foundational for the paradigm. RLHF-style training improves instruction following and reduces harmful behavior, making the policy core more reliable under real user inputs \cite{christiano2017rlhf,ouyang2022instructgpt}. Yet, because agents can take side-effecting actions via tools, safety must be enforced end-to-end across the entire execution graph (retrieval, tool outputs, and action gating), not only in the final response \cite{bai2022constitutional,karpas2022mrkl}.

\subsection{Agent Transformer Definition}
We define an \emph{agent transformer} as a transformer-based policy model embedded in a structured control loop with explicit interfaces to (i) \textbf{observations} from an environment, (ii) \textbf{memory} (short-term working context and long-term state), (iii) \textbf{tools} with typed schemas, and (iv) \textbf{verifiers/critics} that check proposals before side effects occur. The key idea is to make agent behavior a sequence model over \emph{interaction traces}: a trajectory of observations, intermediate thoughts/plans, tool invocations, and outcomes.

% Concretely, an agent transformer can be described by the tuple \((\pi_\theta, \mathcal{M}, \mathcal{T}, \mathcal{V}, \mathcal{E})\), where \(\pi_\theta\) is the transformer policy, \(\mathcal{M}\) is a memory subsystem (retrieval, summaries, state), \(\mathcal{T}\) is a set of tools (APIs, code execution, search, databases), \(\mathcal{V}\) is a set of verifiers/critics, and \(\mathcal{E}\) is the environment. The execution loop repeatedly (a) collects observations from \(\mathcal{E}\), (b) retrieves relevant memory from \(\mathcal{M}\), (c) proposes candidate actions via \(\pi_\theta\), (d) validates them via \(\mathcal{V}\) and tool schemas, and (e) executes selected tool calls in \(\mathcal{T}\) to update \(\mathcal{E}\) and \(\mathcal{M}\).

Concretely, an agent transformer can be described by the tuple
\(\mathcal{A}=(\pi_\theta,\mathcal{M},\mathcal{T},\mathcal{V},\mathcal{E})\),
where \(\pi_\theta\) is the transformer policy, \(\mathcal{M}\) is a memory subsystem
(e.g., retrieval, summaries, and state), \(\mathcal{T}\) is a set of tools (APIs, code execution,
search, databases), \(\mathcal{V}\) is a set of verifiers/critics, and \(\mathcal{E}\) is the environment.
At iteration \(t\), the execution loop proceeds as follows: (i) the agent collects an observation
\(o_t\) from the environment \(\mathcal{E}\); (ii) it retrieves relevant memory \(m_t\) from \(\mathcal{M}\);
(iii) it proposes a candidate action \(a_t\) using the policy \(\pi_\theta\) conditioned on \((o_t,m_t)\);
(iv) it validates \(a_t\) using \(\mathcal{V}\) (and any tool-schema constraints); and (v) it executes the
selected tool call in \(\mathcal{T}\), which updates both the environment \(\mathcal{E}\) and the memory
\(\mathcal{M}\) for the next step.

\begin{equation}
\mathcal{A} \;=\; (\pi_\theta,\mathcal{M},\mathcal{T},\mathcal{V},\mathcal{E}),
\end{equation}
\begin{equation}
o_t \leftarrow \mathrm{Obs}(\mathcal{E}_t), \qquad
m_t \leftarrow \mathrm{Retrieve}(\mathcal{M}_t, o_t),
\end{equation}
\begin{equation}
\tilde{a}_t \sim \pi_\theta(\,\cdot \mid o_t, m_t), \qquad
\hat{a}_t \leftarrow \mathrm{Validate}(\mathcal{V}, \tilde{a}_t),
\end{equation}
\begin{equation}
\mathcal{E}_{t+1} \leftarrow \mathrm{Exec}(\mathcal{E}_t, \mathcal{T}, \hat{a}_t), \qquad
\mathcal{M}_{t+1} \leftarrow \mathrm{Update}(\mathcal{M}_t, o_t, \hat{a}_t, \mathcal{E}_{t+1}).
\end{equation}

% The latest framing is to view this loop as a \emph{risk-aware, budgeted controller}. Not all steps are equal: some actions are reversible (read-only queries) while others are irreversible (writes, deployments, payments). Agent transformers therefore implement decision policies that branch on risk: low-risk actions may run with minimal deliberation, while high-risk actions trigger additional verification, multi-step evidence gathering, or human confirmation \cite{bai2022constitutional,karpas2022mrkl}.
% In this framing, verifiers are not optional add-ons; they define the operational semantics of the agent. A ReAct-style trace is valuable not only for performance but also for governance: it binds decisions to evidence and tool outputs, enabling post-hoc auditing and reproducible replay \cite{yao2023react}. Search-based deliberation (Tree-of-Thoughts) and reflection (Reflexion) can be interpreted as mechanisms for allocating extra compute when uncertainty is high or when failures are detected \cite{yao2023tree,shinn2023reflexion}.

Building on this operational
view, the latest framing is to interpret the loop as a \emph{risk-aware, budgeted controller}:
actions differ in reversibility and potential impact, ranging from low-risk read-only queries to
high-risk irreversible operations such as writes, deployments, or payments. Agent transformers
therefore implement decision policies that branch on risk---low-risk actions may run with minimal
deliberation, while high-risk actions trigger additional verification, multi-step evidence gathering,
or human confirmation \cite{bai2022constitutional,karpas2022mrkl}. In this framing, verifiers are not
optional add-ons but define the operational semantics of the agent: a ReAct-style trace is valuable
not only for performance but also for governance, since it binds decisions to evidence and tool
outputs, enabling post-hoc auditing and reproducible replay \cite{yao2023react}. Likewise, search-based
deliberation (Tree-of-Thoughts) and reflection (Reflexion) can be interpreted as mechanisms for
allocating extra compute when uncertainty is high or when failures are detected \cite{yao2023tree,shinn2023reflexion}.

This abstraction unifies several prominent agent patterns. Retrieval-augmented generation grounds the policy in external evidence by making retrieval a first-class tool and memory operation \cite{lewis2020rag}. ReAct formalizes the interleaving of reasoning and acting by alternating between deliberation tokens and tool calls, improving grounding and enabling evidence-backed traces \cite{yao2023react}. MRKL-style systems route tasks to specialized tools, separating language understanding from deterministic components and improving governability \cite{karpas2022mrkl}. Critic/reflection mechanisms (e.g., Reflexion) add an internal feedback channel that reduces compounding errors and supports iterative repair \cite{shinn2023reflexion}. Search-based deliberation (Tree-of-Thoughts) treats planning as exploring a space of action candidates, trading compute for reliability \cite{yao2023tree}. Finally, multi-agent frameworks implement the same abstraction with multiple policies that communicate via messages, enabling specialization and cross-checking at the cost of coordination complexity \cite{wu2023autogen,li2023camel}.

\subsection{Agent Transformer Creation}

\begin{figure}[thbp]
\centering
\includegraphics[width=0.8\linewidth]{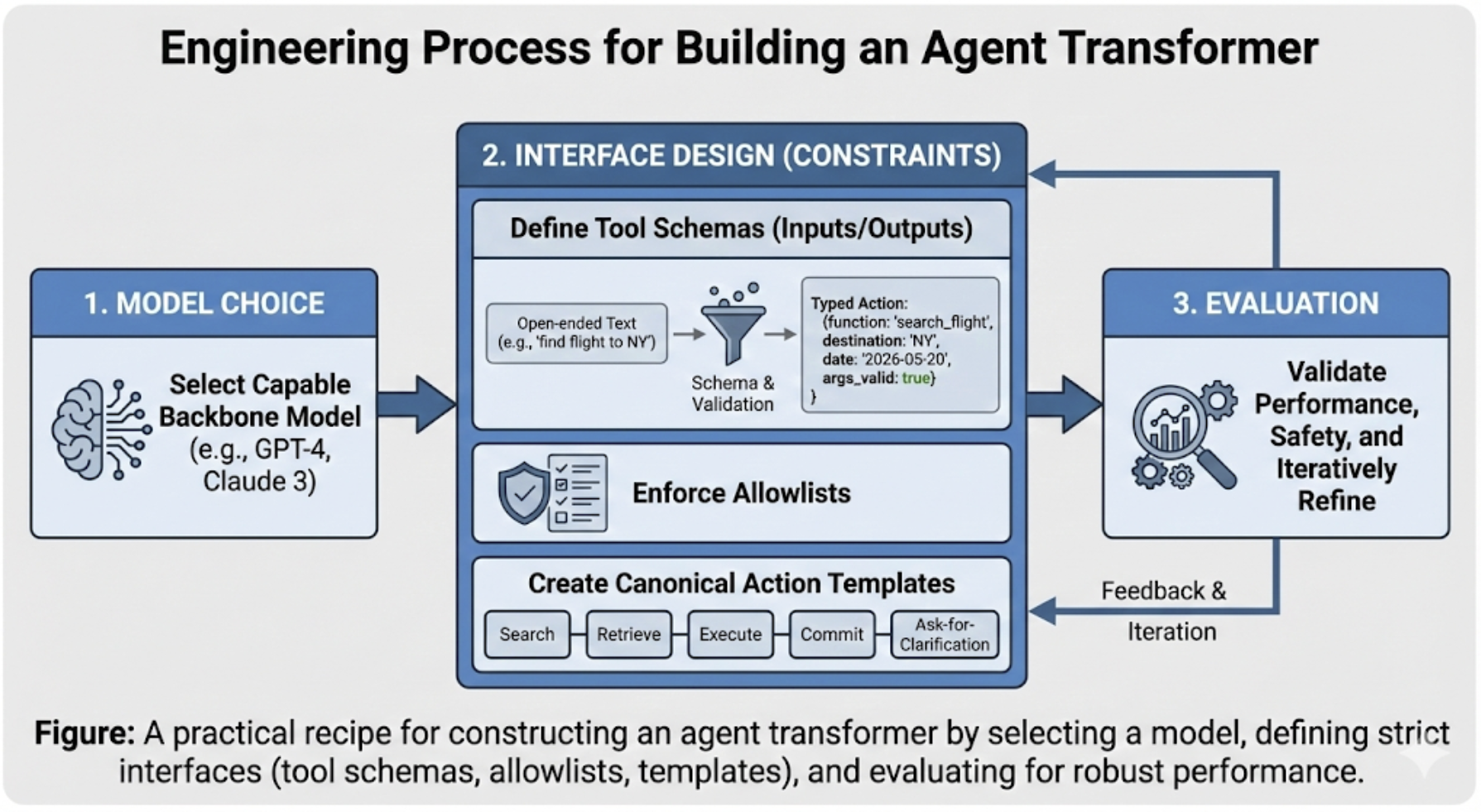}
\caption{Agent transformer abstraction with explicit interfaces to memory, tools, verifiers, and environment}
\label{fig:section3}
\end{figure}

Fig.~\ref{fig:section3} illustrates the agent transformer abstraction, emphasizing explicit interfaces to memory, tools, verifiers, and the environment.

Building an agent transformer in practice is an engineering process that combines model choice, interface design, and evaluation. A common recipe starts by selecting a capable backbone model and then constraining it through \emph{interfaces}: define tool schemas (inputs/outputs), enforce allowlists, and create a small set of canonical action templates (search, retrieve, execute, commit, ask-for-clarification). Tool schemas reduce brittleness by turning open-ended text into typed actions, and they enable automatic argument validation before execution \cite{schick2023toolformer,karpas2022mrkl}.

Next, design the control loop. A minimal loop is (retrieve context) \(\rightarrow\) (plan) \(\rightarrow\) (act via tools) \(\rightarrow\) (verify) \(\rightarrow\) (update memory) \(\rightarrow\) (repeat), which is closely aligned with ReAct and reflection patterns \cite{yao2023react,shinn2023reflexion}. For harder tasks, add deliberation depth: tree-style search over candidate actions, self-consistency reruns, and explicit critics that check for policy violations, missing evidence, or unsafe side effects \cite{yao2023tree,wang2022selfconsistency,bai2022constitutional}. In tool-rich environments (web, code, enterprise systems), the most important design choice is often \emph{when to allow side effects}: high-impact actions should require stronger verification, human confirmation, or sandboxed execution.

Then, choose learning signals that match the environment. In many deployments, supervised finetuning on traces (tool calls + outcomes) provides strong initial behavior; preference optimization and RLHF improve instruction following and refusal behavior under adversarial prompts \cite{ouyang2022instructgpt,rafailov2023dpo,christiano2017rlhf}. Tool-use learning can be bootstrapped from synthetic traces or self-supervision (Toolformer-style), reducing the need for brittle prompt engineering \cite{schick2023toolformer}. For embodied or real-time control layers, combine an LLM planner with specialized controllers trained by RL/IL to satisfy timing and safety constraints \cite{driess2023palme,brohan2023rt2}.

Recent practice emphasizes a \emph{trace-first data flywheel}: run the agent in realistic environments, log full trajectories (prompts, tool calls, tool outputs, and outcomes), and continuously mine failures for targeted improvements (better prompts, new tools, better verifiers, or finetuning on corrected traces). This shifts learning from one-off model training to continuous system refinement, and it makes evaluation suites and regression tests essential engineering artifacts \cite{qin2023toolbench,liu2023agentbench}.
Another emerging best practice is to explicitly separate \emph{planning} from \emph{execution}: a planner proposes a plan with explicit constraints and success criteria, while an executor carries out the plan under stricter tool permissions and validation. This separation improves controllability, supports human-in-the-loop approval for high-impact steps, and reduces the blast radius of failures \cite{karpas2022mrkl,bai2022constitutional}.
Finally, deployment increasingly depends on operational discipline: caching and summarization to control context growth, sandboxing for code and web actions, and policy-as-code gates for compliance. These choices do not merely improve engineering robustness; they change the effective agent policy by constraining what information and actions are available under budget and safety constraints \cite{yao2023react,bai2022constitutional}.

Finally, evaluate the agent transformer as a system, not a model. Benchmarks that stress realistic tool use and long-horizon execution reveal failures that do not appear in static QA: web task suites, software engineering issue resolution, and tool-use benchmarks quantify end-to-end reliability and tool correctness \cite{zhou2023webarena,jimenez2023swebench,qin2023toolbench,liu2023agentbench}. Report not only success rate but also cost/latency, trace completeness, robustness under variability, and safety violations, because these determine whether an agent is deployable under real constraints \cite{bai2022constitutional}.

\section{Agent AI Learning}
\label{sec:learning}

\begin{figure}[thbp]
\centering
\includegraphics[width=0.8\linewidth]{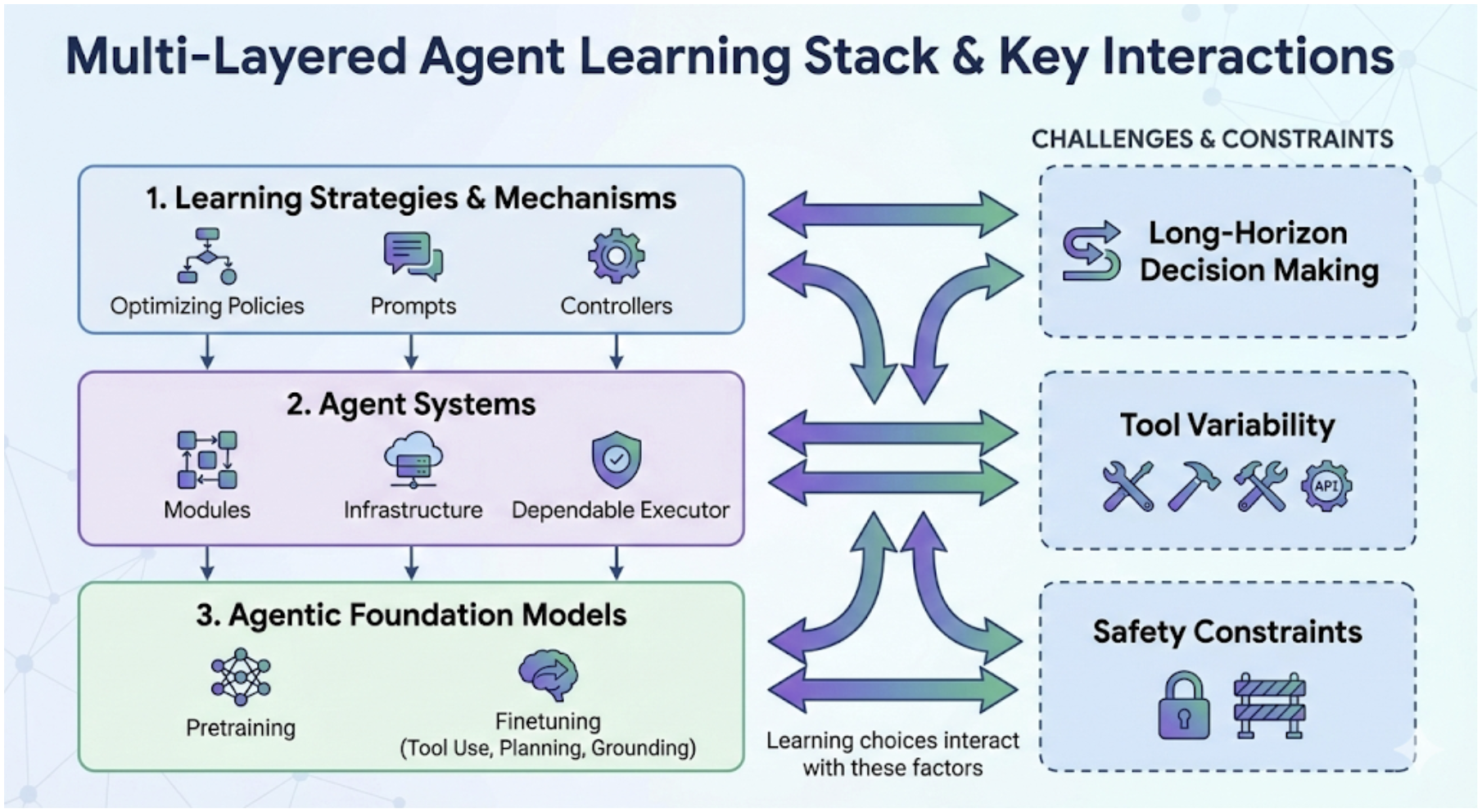}
\caption{Overview of agent AI learning across mechanisms, systems, and foundation models}
\label{fig:section4}
\end{figure}

Agent learning spans multiple layers of the stack: (i) \emph{learning strategies and mechanisms} (how policies, prompts, and controllers are optimized), (ii) \emph{agent systems} (how modules and infrastructure turn a model into a dependable executor), and (iii) \emph{agentic foundation models} (how pretraining and finetuning shape tool use, planning, and grounding). This section emphasizes how learning choices interact with long-horizon decision making, tool variability, and safety constraints \cite{luo2025llmagentsurvey,sang2025beyondpipelines}. Fig.~\ref{fig:section4} outlines the learning stack for agentic systems, connecting mechanisms, system-level engineering, and foundation-model adaptation.

\subsection{Strategy and Mechanism}
\subsubsection{Reinforcement Learning (RL)}
RL is a natural fit for agentic behavior because it directly optimizes long-horizon returns under interaction, typically formalized as a Markov decision process with a policy that maximizes expected discounted reward \cite{sutton2018rlbook,puterman1994mdp}.
For agents, the appeal is that RL optimizes \emph{behavior} rather than one-step prediction: it can learn when to gather information, when to act, and how to recover from errors across multi-step trajectories.

Algorithmically, modern deep RL spans value-based learning (e.g., Q-learning variants) and policy-gradient methods, which differ in stability, sample efficiency, and how they handle continuous actions \cite{mnih2015dqn,schulman2015trpo,schulman2017ppo,haarnoja2018sac}.
Hierarchical RL (e.g., options) is especially relevant to agents because it provides a learning substrate for reusable skills, temporal abstraction, and planner--controller decompositions \cite{dayan1993options}.
In embodied settings, RL often operates at the low-level control layer where timing constraints are strict and simulation can provide abundant interaction, while higher-level reasoning and language grounding are handled by LLMs or planners \cite{brohan2023rt2,driess2023palme}.

\begin{figure}[thbp]
\centering
\includegraphics[width=0.8\linewidth]{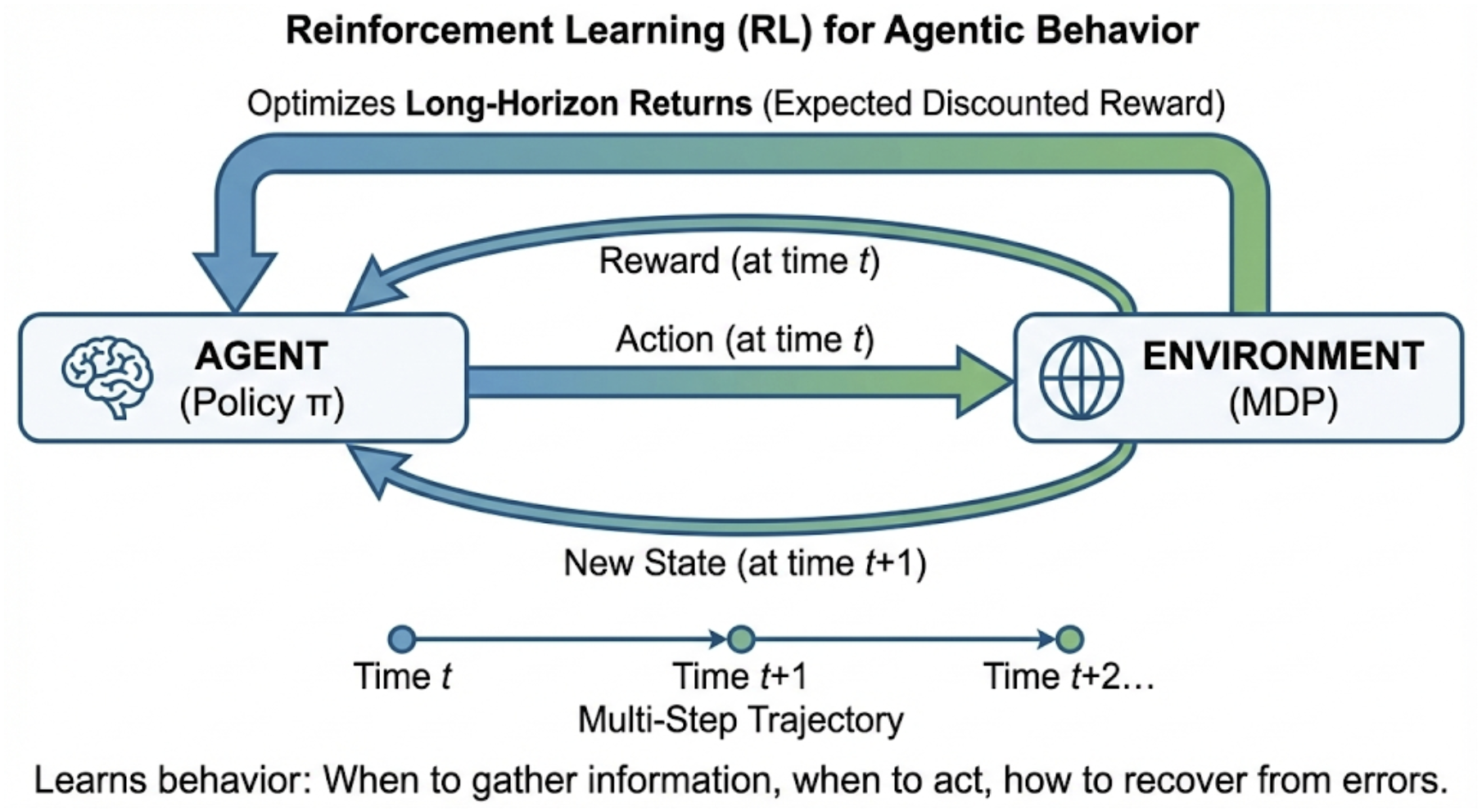}
\caption{Reinforcement learning (RL) pipeline for agent policies and controllers}
\label{fig:section5}
\end{figure}

Fig.~\ref{fig:section5} provides a schematic view of how reinforcement learning fits into agentic decision making and control.

Empirically, classic successes in competitive games highlight the benefits of end-to-end optimization and disciplined interfaces, yielding strong execution and stability under distributional variability induced by opponents \cite{berner2019openaifive,vinyals2019alphastar}.
However, RL in tool-rich real-world settings faces bottlenecks that are specific to agent workflows: sparse or delayed rewards, expensive rollouts (because tool calls and environment steps are costly), and safety constraints that limit exploration.
This motivates safer and more data-efficient regimes such as offline RL and constrained/safe RL, where the agent is optimized from logged trajectories and policy constraints bound undesirable actions \cite{sutton2018rlbook,kumar2020cql,fujimoto2019bcq,achiam2017constrained,garcia2015safe}.

In LLM-centric agents, RL also appears in \emph{alignment} and \emph{preference optimization}: RL from human feedback shapes response behavior and instruction following \cite{christiano2017rlhf,ouyang2022instructgpt}, while constitution-style policy feedback and direct preference optimization provide alternative alignment mechanisms that are often easier to operationalize \cite{bai2022constitutional,rafailov2023dpo}. Variants such as ``verbal reinforcement'' (reflection-based self-improvement) adapt the idea of learning from feedback to language-agent loops \cite{shinn2023reflexion}.
Overall, RL is most effective when (i) the environment and action space are sufficiently well-defined, (ii) interaction can be scaled (simulation or robust logging), and (iii) safety and evaluation are treated as system properties rather than post-hoc filters.

\subsubsection{Imitation Learning (IL)}

\begin{figure}[thbp]
\centering
\includegraphics[width=0.8\linewidth]{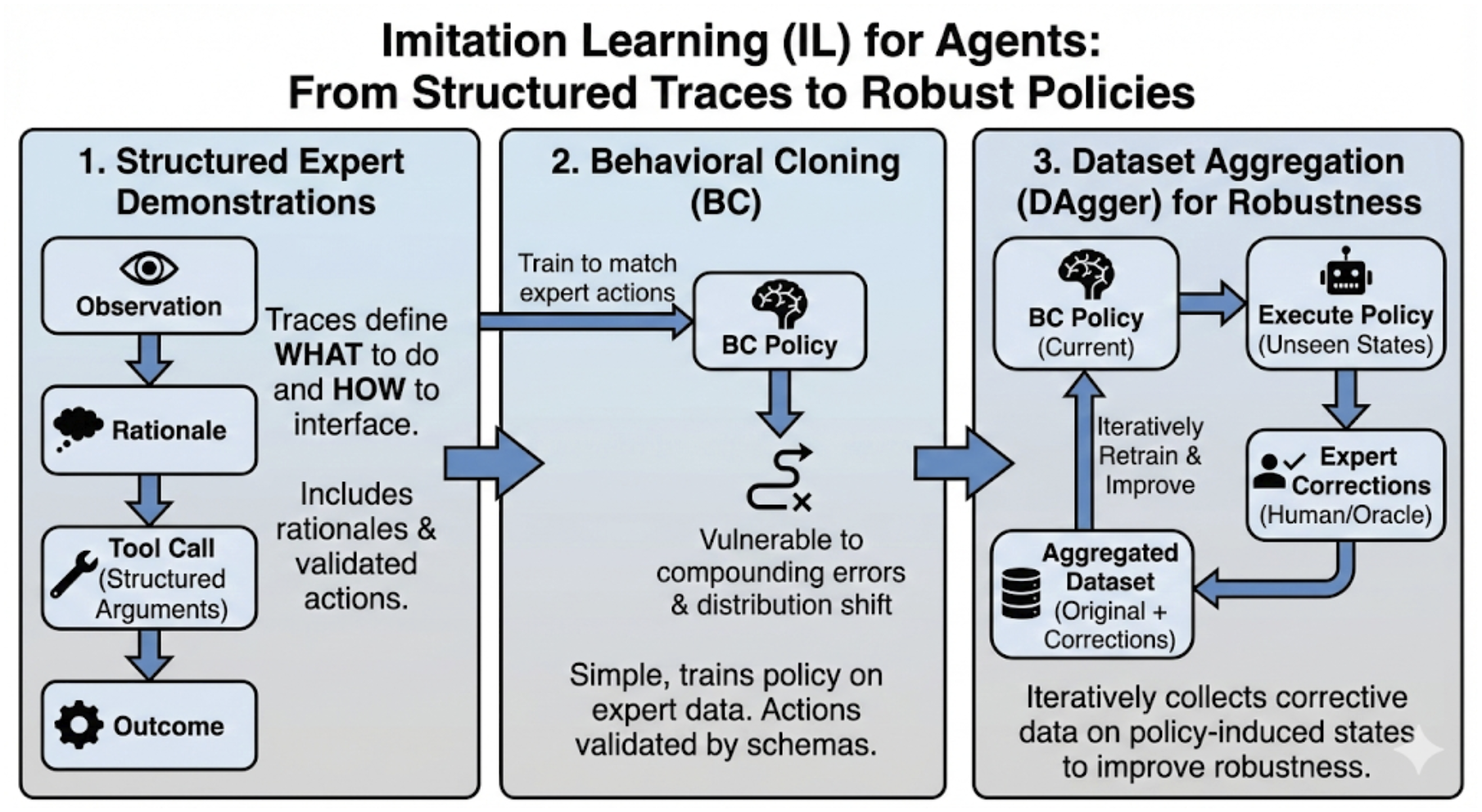}
\caption{Imitation learning (IL) from demonstrations and interaction traces}
\label{fig:section6}
\end{figure}

IL provides a pragmatic route to competent behavior when expert demonstrations (human traces, scripted policies, or curated tool trajectories) are available.
For agents, demonstrations are often \emph{structured traces}: sequences of observations, intermediate rationales, tool calls, and outcomes that define not only \emph{what} to do but also \emph{how} to interface with tools reliably. Fig.~\ref{fig:section6} summarizes the imitation learning pipeline for acquiring agent behaviors from demonstrations and interaction traces.

The simplest form, behavioral cloning, trains a policy to match expert actions, which is attractive for tool calling because actions can be represented as structured arguments and validated by schemas \cite{pomerleau1991alvinn}.
But pure cloning is vulnerable to compounding errors: small deviations from the expert distribution can lead to unseen states where the policy has no guidance.
Dataset aggregation methods such as DAgger address this by iteratively collecting corrective demonstrations on states induced by the learned policy, improving robustness under distribution shift \cite{ross2011dagger}.

Beyond direct imitation, inverse RL and adversarial imitation aim to infer objectives or match expert occupancy measures.
For example, GAIL learns policies by matching expert behavior distributions without explicitly specifying rewards, which can be useful when the ``reward'' is implicit in expert traces (e.g., how humans navigate UIs or debug code) \cite{ho2016gail}.
In agent deployments, IL is often the dominant learning signal when high-quality traces exist (enterprise workflows, customer-support playbooks, or curated code-repair sequences), because it avoids unsafe exploration and is cheaper than RL in tool-rich environments.

However, IL inherits biases and coverage gaps from the demonstration set: experts may follow organization-specific habits, may omit edge cases, and may not represent adversarial inputs (prompt injection, malicious pages, or ambiguous requests).
Therefore, IL-trained behaviors often benefit from verification and repair loops (critics, self-correction, constrained execution) to handle out-of-distribution cases and to avoid blindly copying brittle heuristics \cite{yao2023react,shinn2023reflexion}.
In practice, robust agent learning combines IL for baseline competence with system-level guardrails and post-hoc verification for reliability under long-horizon compounding errors.

\subsubsection{Traditional RGB}

\begin{figure}[thbp]
\centering
\includegraphics[width=0.8\linewidth]{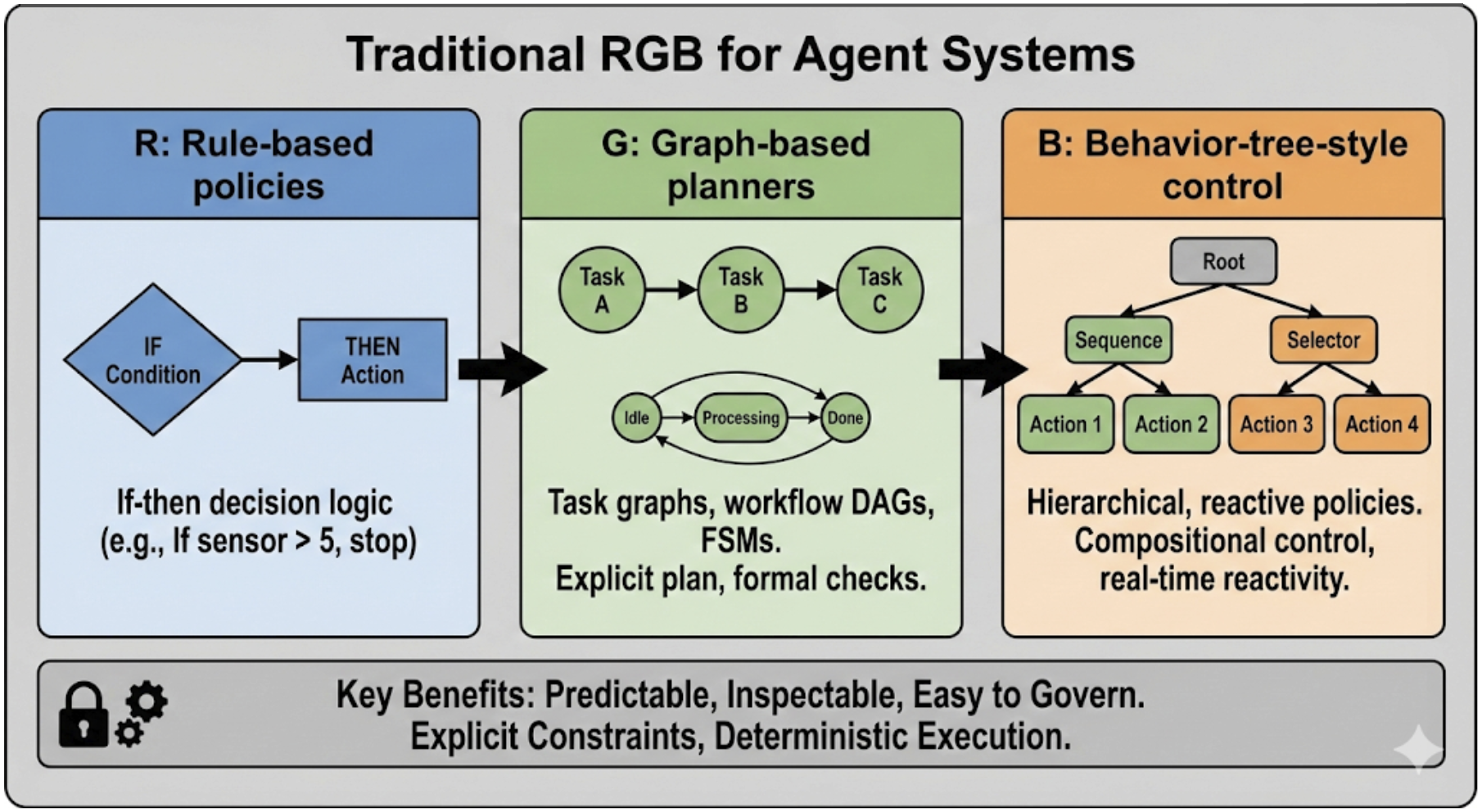}
\caption{Traditional RGB components: rule-based policies, graph planners, and behavior trees}
\label{fig:section7}
\end{figure}

Fig.~\ref{fig:section7} highlights traditional rule/graph/behavior-tree (RGB) components that remain important baselines and safety interfaces in agent systems.
Before LLM-centric agents, many production systems relied on \emph{Traditional RGB} components: \textbf{R}ule-based policies (if--then decision logic), \textbf{G}raph-based planners (task graphs, workflow DAGs, FSMs), and \textbf{B}ehavior-tree-style control (hierarchical, reactive policies).
These approaches remain useful because they are predictable, inspectable, and easy to govern: constraints can be encoded explicitly, and execution can be audited deterministically.

Behavior trees in particular provide compositional control and real-time reactivity, making them attractive for robotics and games where strict timing and safety envelopes must be enforced \cite{colledanchise2018behaviortrees}.
Graph-based planning subsumes workflow and task-graph execution: the ``plan'' is explicit (nodes/edges), enabling formal checks (reachability, preconditions, approval gates) and deterministic replay, which is essential for compliance and debugging.

The main limitation of RGB approaches is brittleness: handcrafted rules may not generalize, and enumerating all edge cases becomes infeasible as environments become open-ended.
This is where LLM-centric components provide value: they can interpret ambiguous instructions, propose candidate plans, and adapt to novel inputs.
Modern stacks therefore hybridize: LLMs propose goals, explanations, or candidate actions, while RGB components enforce safety, timing, and domain rules (e.g., cooldowns in games, approvals in enterprise workflows, or safety envelopes in robotics) \cite{karpas2022mrkl,ahn2022saycan}.

From a learning perspective, RGB components also function as \emph{inductive biases} and \emph{interfaces}: they constrain the action space and reduce the burden on statistical learning, often improving reliability and making evaluation more reproducible.
In many production deployments, the highest-value learning work is not replacing RGB entirely, but learning \emph{better routing and parameterization} of RGB modules (which rule to apply, which graph branch to follow, which behavior-tree subtree to activate) under contextual signals from LLMs and tools \cite{yao2023react}.

\subsubsection{In-context Learning}

\begin{figure}[thbp]
\centering
\includegraphics[width=0.8\linewidth]{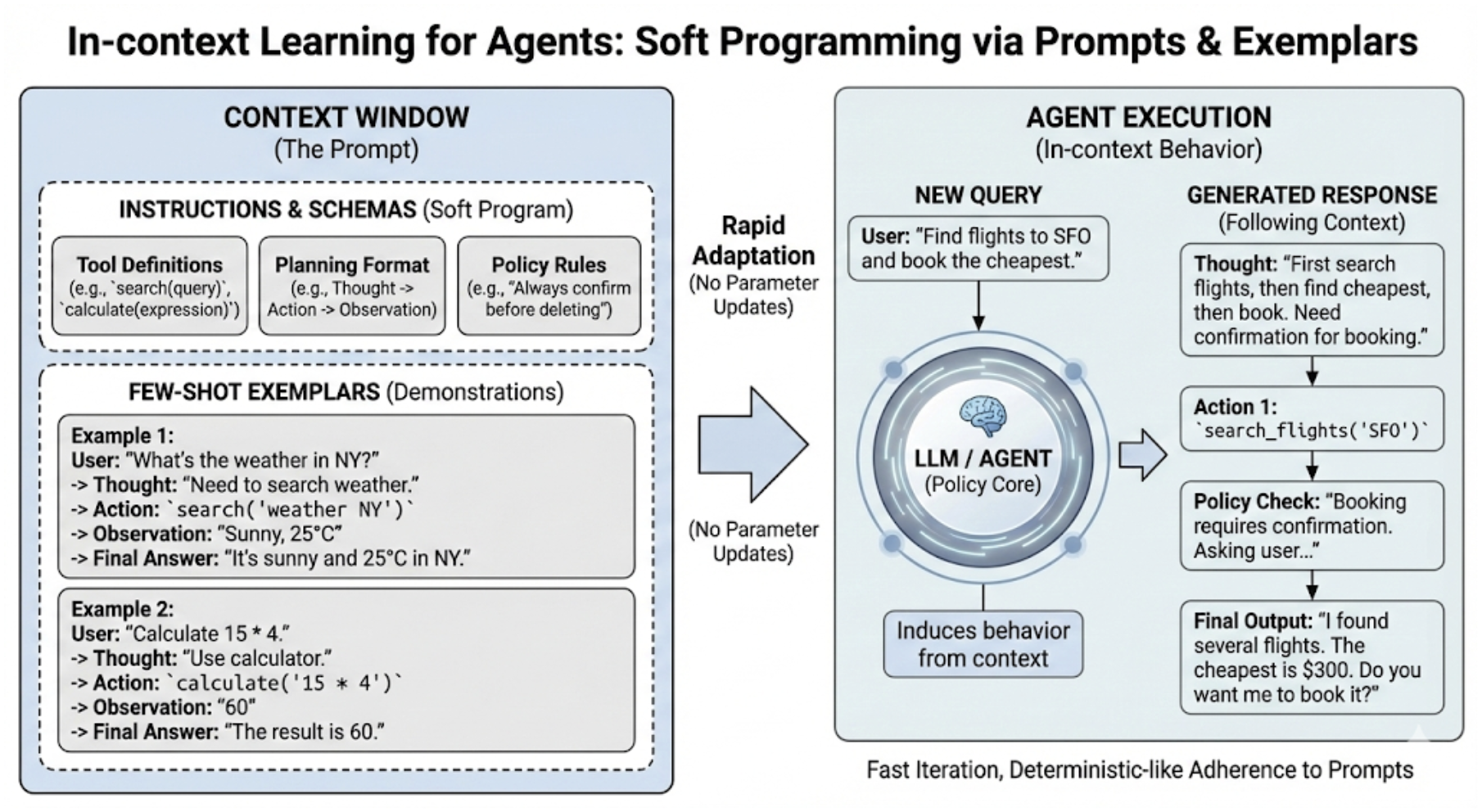}
\caption{In-context learning for agents via prompts, exemplars, and action schemas}
\label{fig:section8}
\end{figure}

In-context learning enables rapid task adaptation via prompting and exemplars without parameter updates. Fig.~\ref{fig:section8} depicts in-context learning as a practical mechanism for teaching agent protocols (formats, schemas, and interaction patterns) without parameter updates. For agents, this includes demonstrations of tool schemas, planning formats, and interaction protocols (e.g., how to interleave reasoning and actions).
Empirically, large language models exhibit strong in-context learning capabilities where few-shot exemplars induce task behavior without finetuning \cite{brown2020gpt3,min2022rethinking}.
For agent settings, in-context learning acts as ``soft programming'': prompts define action formats, tool-call schemas, and policies (what is allowed, what requires confirmation), enabling fast iteration without retraining.

Reasoning-augmented prompting is a key enabler. Chain-of-thought prompting improves multi-step reasoning and decomposition, which directly translates to better planning and tool selection in agents \cite{wei2022cot,kojima2022zeroshotcot,wang2022selfconsistency}.
ReAct-style prompting operationalizes in-context learning by binding reasoning to tool use, improving grounding and making intermediate decisions inspectable and auditable \cite{yao2023react}.
Self-consistency and related sampling-based methods further stabilize in-context behaviors by aggregating multiple reasoning paths, which is especially useful under nondeterminism and partial observability \cite{wang2022selfconsistency}.

However, in-context learning has well-known system-level failure modes for agents: context growth increases cost/latency, long prompts can dilute critical constraints, and retrieved text can introduce prompt-injection instructions that override policy.
Therefore, in-context learning is most effective when paired with memory (summaries, persistent state), retrieval grounded in trusted sources, and strict tool interfaces that cannot be bypassed by text alone \cite{lewis2020rag,bai2022constitutional}.
In production, prompts are often treated as versioned artifacts with evaluation suites and regression checks, because small changes in examples or formatting can shift behavior dramatically.

Finally, in-context learning interacts strongly with tool ecosystems: exposing a tool via a schema and exemplars is a kind of ``few-shot tool learning,'' which can substitute for explicit finetuning when new tools are added frequently \cite{schick2023toolformer,karpas2022mrkl}.
This emphasizes a design principle: if tool interfaces are stable and well-specified, in-context learning can deliver rapid capability gains with relatively predictable governance and rollback.

\subsubsection{Optimization in the Agent System}
Agent performance is a systems optimization problem as much as a modeling problem: reliability, latency, and cost are shaped by orchestration policies (how many calls, which tools, how much verification, and when to backtrack).
Search-based planning (e.g., exploring alternative action sequences) improves hard tasks where a single rollout is unreliable, but increases compute and requires careful termination criteria and evaluation functions \cite{yao2023tree}.
In LLM agents, search often operates over \emph{action proposals} rather than over low-level states: the agent samples candidate tool calls or subplans, scores them (via critics, heuristics, or tool-based checks), and commits to the best candidate.

\begin{figure}[thbp]
\centering
\includegraphics[width=0.8\linewidth]{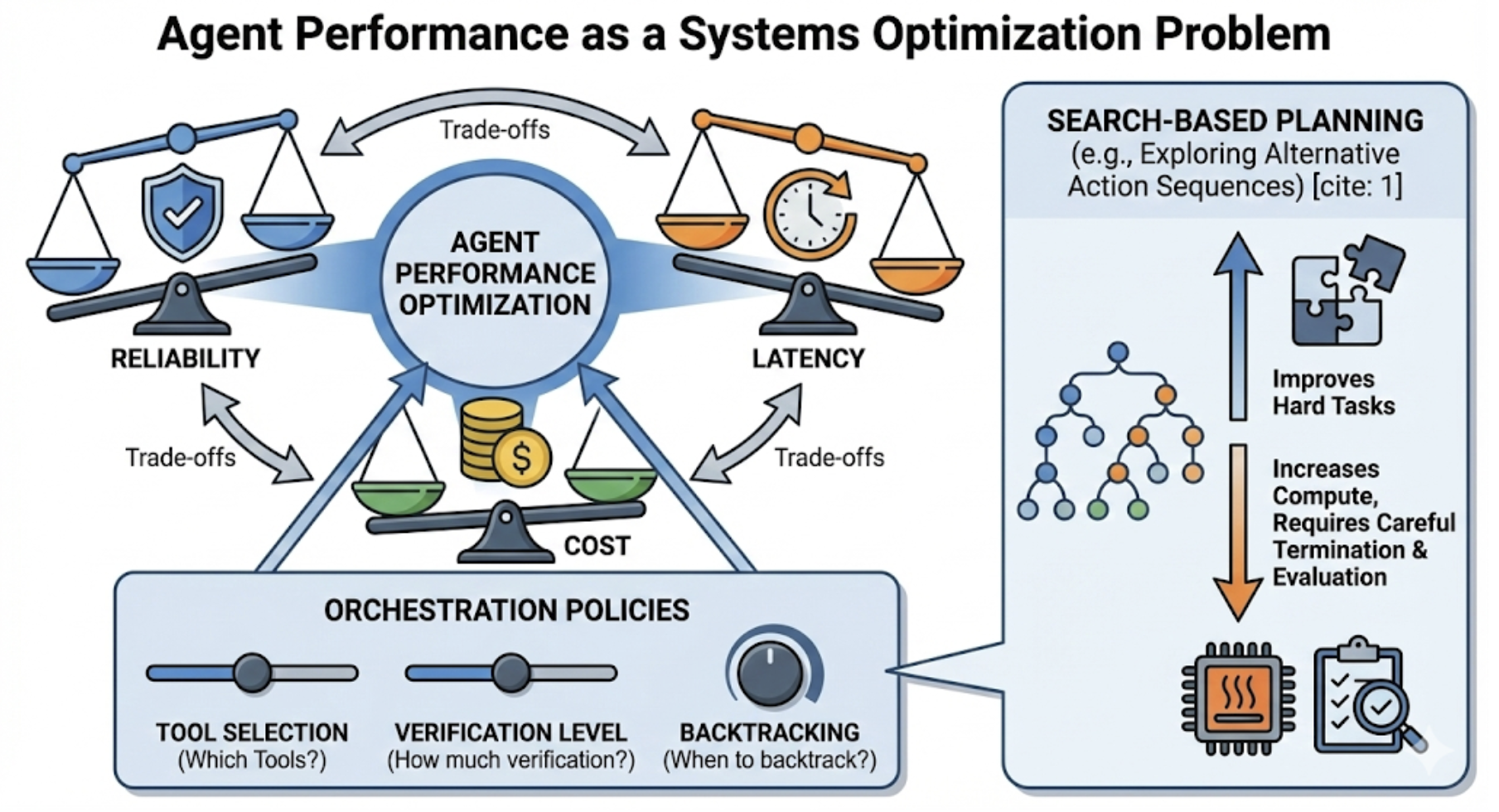}
\caption{Optimization Problem for AI Agent}
\label{fig:section31}
\end{figure}

Fig.~\ref{fig:section31} provides an optimization-oriented view of agent design, highlighting the core trade-offs between reliability, latency, and cost under constrained orchestration policies.

Verification and reflection loops trade additional computation for lower failure rates by checking actions and revising plans in response to tool outputs or detected inconsistencies \cite{shinn2023reflexion,yao2023react}.
This is closely related to ``test-time compute'' scaling: rather than increasing model size, systems increase the number of controlled deliberation steps (reranking, self-consistency, backtracking).
In practice, verification is most valuable for high-impact or irreversible actions (writing to a database, merging code, sending messages), where the cost of a mistake dominates latency concerns \cite{karpas2022mrkl,bai2022constitutional}.

Optimization also includes purely systems concerns: caching frequent retrievals, batching model calls, using smaller models for routing/moderation, and compressing memory via summaries to control context growth.
These choices change the agent's effective policy by altering what context is available and which tools are feasible under latency/cost constraints.
Therefore, optimization must be evaluated end-to-end (success rate, safety, and cost/latency) rather than at the component level \cite{qin2023toolbench,liu2023agentbench}.

Practical deployments adopt adaptive optimization: fast-path execution for routine cases, slower verified paths for high-risk actions, explicit budgets (time, tokens, tool calls), and permission gates that bound side effects even when the model is capable \cite{karpas2022mrkl,bai2022constitutional}.
This perspective treats agent learning as a \emph{co-design} problem between models and orchestration, where the most impactful improvements often come from changing the decision loop rather than changing the base model.

\subsection{Agent Systems}
\subsubsection{Agent Modules}

\iffalse
\begin{figure}[thbp]
\centering
\includegraphics[width=0.8\linewidth]{section9.png}
\caption{Core agent modules: policy core, memory, planners, tool routers, and verifiers}
\label{fig:section9}
\end{figure}
\fi

At the system level, ``learning'' includes how the agent is modularized into components with clear contracts. Common modules include an LLM policy core, retrieval/memory, planners, tool routers, and critics/verifiers.
MRKL-style routing separates language understanding from specialized tools, improving controllability and allowing toolchains to evolve without retraining the core model \cite{karpas2022mrkl}.
ReAct-style execution couples reasoning with actions to produce traceable trajectories, while reflection mechanisms provide a structured way to recover from errors and reduce compounding failures \cite{yao2023react,shinn2023reflexion}.

From a learning viewpoint, modularization changes what must be learned by the core model. For example, if retrieval is delegated to a tool, the model learns \emph{query formulation} rather than memorizing facts; if a verifier exists, the model can learn to propose candidates and rely on checks to reject unsafe actions.
This enables a division of labor where smaller or specialized models can handle routing, safety classification, or summarization, reducing cost and improving predictability \cite{karpas2022mrkl,openai2023gpt4}.

Memory modules are particularly central for long-horizon agents: episodic memory (what happened), semantic memory (facts), and procedural memory (skills) support coherence beyond raw context windows.
Retrieval-augmented generation provides a standard mechanism for grounding and reduces hallucinations by binding claims to retrieved evidence \cite{lewis2020rag}.
However, memory also increases attack surface (retrieved prompt injection) and introduces consistency challenges when stored state conflicts with new observations; critics/verifiers mitigate this by checking claims against tool outputs and trusted sources \cite{bai2022constitutional,yao2023react}.
Experience-oriented agent architectures further motivate richer long-term memory (personas, episodic summaries, skills) to support coherence and learning over extended interaction \cite{park2023generativeagents,wang2023voyager}.

Multi-agent variants decompose tasks across roles (planner, executor, reviewer), enabling cross-checking and specialization, but they introduce coordination costs (latency, token usage) and can amplify inconsistency when agents disagree \cite{wu2023autogen,li2023camel}.
In practice, role separation works best when roles have distinct tool access and explicit handoff artifacts (plans, checklists, traces), so that disagreements can be resolved via evidence rather than free-form debate.

\subsubsection{Agent Infrastructure}
\begin{figure}[thbp]
\centering
\includegraphics[width=\linewidth]{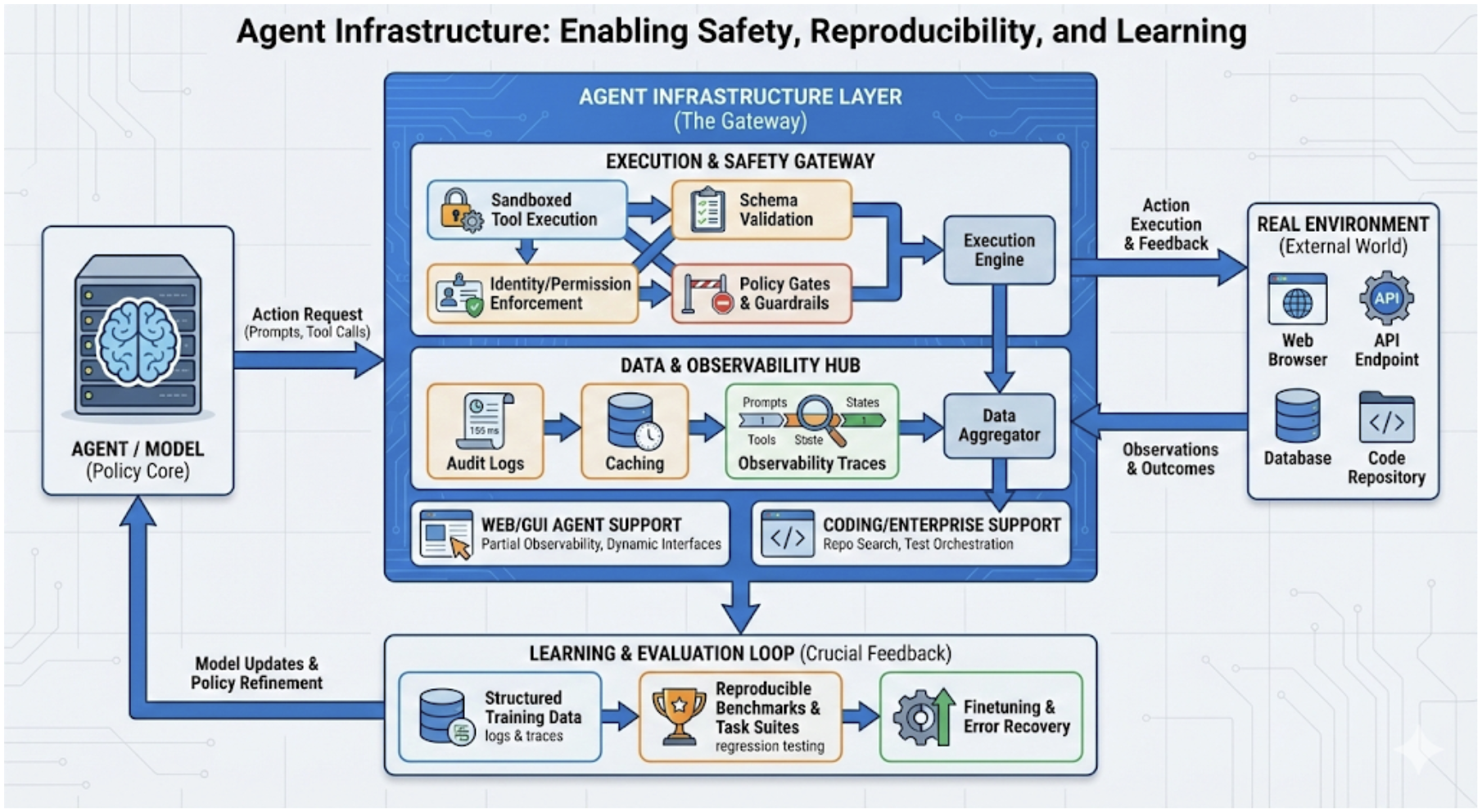}
\caption{Agent infrastructure for safe deployment: sandboxing, schemas, permissions, and logging}
\label{fig:section10}
\end{figure}

Infrastructure determines whether an agent can operate safely and reproducibly in real environments. Key elements include sandboxed tool execution, schema validation, identity/permission enforcement, audit logs, caching, and observability (traces of prompts, tool calls, and intermediate states). Fig.~\ref{fig:section10} sketches the infrastructure layer (sandboxing, schemas, permissions, logging) required for safe and reproducible agent deployment.

For web and GUI agents, infrastructure must handle partial observability and dynamic interfaces; standardized environments and benchmarks improve comparability and reveal brittleness undrealistic variability \cite{zhou2023webarena}.
For coding agents and enterprise workflows, infrastructure often includes repository search, test orchestration, and policy gates; end-to-end benchmarks highlight failures that only appear when tools, environments, and long-horizon dependencies interact \cite{jimenez2023swebench,liu2023agentbench,qin2023toolbench}.
Across domains, guardrails must be enforced end-to-end (including tool outputs and retrieved text) to mitigate prompt injection and unsafe side effects \cite{bai2022constitutional}.

Crucially, infrastructure determines \emph{what is learnable}. If tool calls are logged with arguments and outcomes, they become training data for tool-use finetuning and error recovery; if traces are missing or privacy-restricted, learning must rely on weaker signals.
Similarly, schema validation and sandboxing turn an open-ended ``action'' into a constrained interface, improving reliability and reducing catastrophic failures when models hallucinate tool arguments \cite{schick2023toolformer,karpas2022mrkl}.

Evaluation infrastructure is part of learning: reproducible benchmarks and task suites enable regression testing and ablations over prompts, routing, memory policies, and verification depth.
For interactive environments, standardized suites make it possible to report robustness to variability (layout changes, tool failures) and to quantify tool-use correctness beyond superficial answer quality \cite{zhou2023webarena,qin2023toolbench,liu2023agentbench}.
In enterprise and safety-critical deployments, auditability (who did what, when, with what evidence) is a core requirement; therefore, learning objectives must include not only task success but also trace completeness and policy compliance \cite{bai2022constitutional,karpas2022mrkl}.

\subsection{Agentic Foundation Models (pretraining and finetune level)}

\begin{figure}[thbp]
\centering
\includegraphics[width=0.8\linewidth]{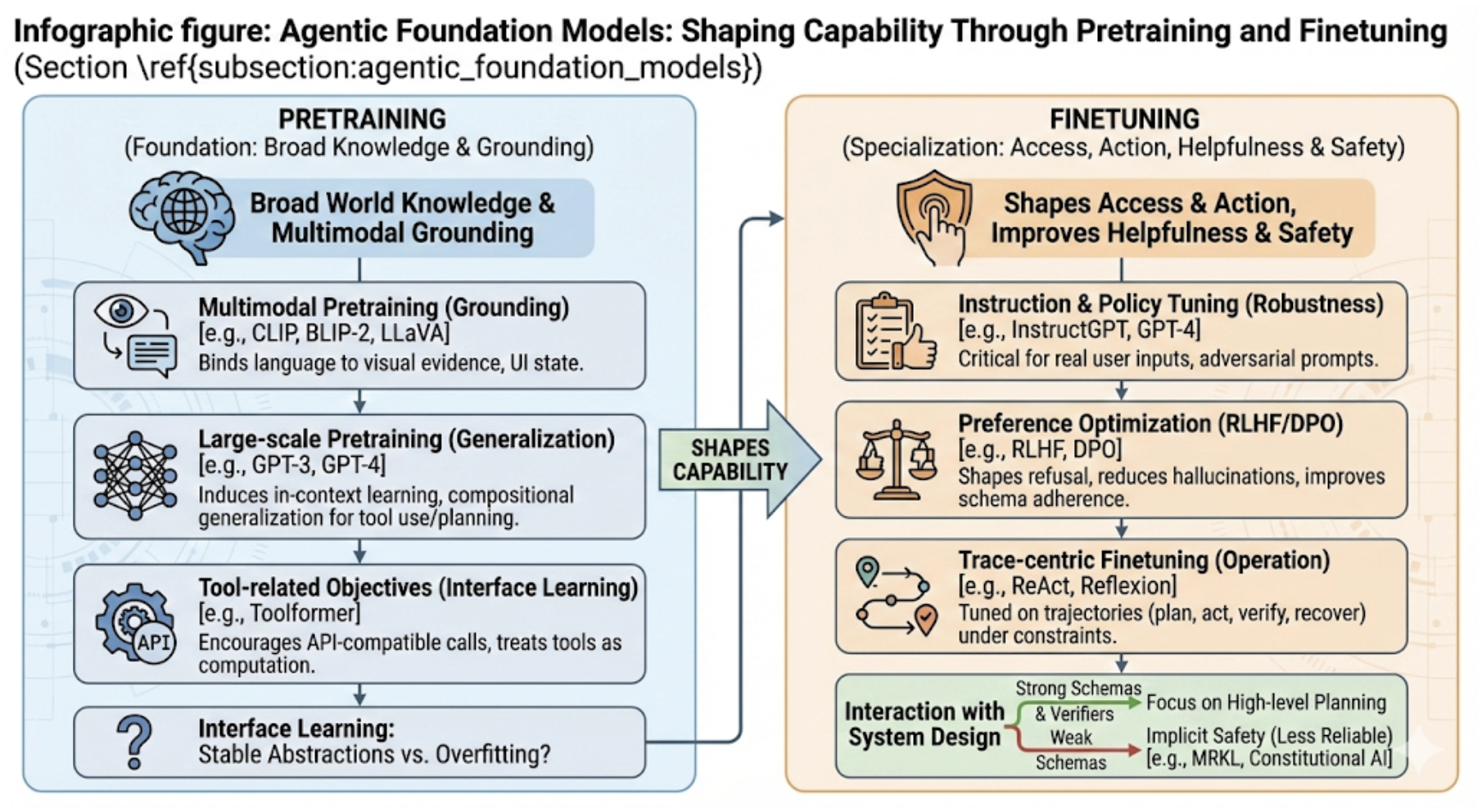}
\caption{Agentic foundation models: pretraining and finetuning for tool use and planning}
\label{fig:section11}
\end{figure}

Foundation models shape agent capability through both representation learning and alignment. Pretraining builds broad world knowledge and multimodal grounding, while finetuning (instruction/policy tuning, tool-use tuning) shapes how that knowledge is accessed and acted upon. Fig.~\ref{fig:section11} summarizes how pretraining and finetuning choices shape tool use, planning, and grounding in agentic foundation models.

Pretraining: Multimodal pretraining improves grounding and perception, enabling agents to bind language to visual evidence and UI state \cite{radford2021clip,li2023blip2,liu2023llava}.
More broadly, large-scale pretraining induces in-context learning and compositional generalization, which are prerequisites for tool use and long-horizon planning in open-ended environments \cite{brown2020gpt3,openai2023gpt4}.
Tool-related pretraining objectives can explicitly encourage models to emit API-compatible calls and to treat tools as part of the computation graph rather than as post-hoc decorations \cite{schick2023toolformer}.
For agentic behavior, the key pretraining question is \emph{interface learning}: does the model learn stable abstractions for actions, observations, and constraints, or does it overfit to surface patterns that break under distribution shift?

Finetuning: Instruction and policy tuning improve helpfulness and safety, and they are critical for making agents robust under real user inputs and adversarial prompts \cite{ouyang2022instructgpt,bai2022constitutional,openai2023gpt4}.
Many systems combine supervised instruction tuning with preference-based optimization (RLHF or direct preference optimization) to shape refusal behavior, reduce hallucinations, and improve adherence to tool schemas \cite{christiano2017rlhf,ouyang2022instructgpt,rafailov2023dpo}.
Agentic finetuning is increasingly trace-centric: models are tuned on trajectories that include tool calls, intermediate checks, and corrected failures, so that the model learns not only to answer but also to \emph{operate}---plan, act, verify, and recover---under constraints \cite{yao2023react,shinn2023reflexion}.
Finally, finetuning interacts with system design: if orchestration enforces strict schemas and verifiers, finetuning can focus on high-level planning and query formulation; if schemas are weak, finetuning must implicitly learn safety and interface constraints, which is less reliable and harder to audit \cite{karpas2022mrkl,bai2022constitutional}.

\section{Agent AI Taxonomy}
\label{sec:taxonomy}

\iffalse
\begin{figure}[thbp]
\centering
\includegraphics[width=0.8\linewidth]{section12.png}
\caption{Overview of the agent AI taxonomy used in this survey}
\label{fig:section12}
\end{figure}
\fi

To connect architectural choices to deployment requirements, we categorize agent systems by the \emph{dominant locus of interaction} (text/tools, physical embodiment, simulated environments), the \emph{generative target} (content/worlds/experiences), and the \emph{reasoning substrate} (knowledge, logic, emotion, neuro-symbolic structure). This taxonomy is intended to be pragmatic: categories reflect the constraints that most strongly shape system design (observability, safety, latency, verification, and evaluation) \cite{sapkota2025agentsvsagentic,zhou2024taxonomyarchexoptions,aegis2025agentenvfailures}.

\subsection{Generalist Agent Areas}
Generalist agents aim to solve heterogeneous tasks across domains (coding, browsing, analytics, and enterprise workflows) using a shared policy core plus modular tools and memory \cite{openai2023gpt4,brown2020gpt3}.
\textbf{Challenges.} The dominant failure mode is long-horizon compounding error under tool and environment variability: the agent must retrieve the right context, choose correct tools and arguments, and recover from partial failures (timeouts, flaky tests, UI changes) \cite{yao2023react,qin2023toolbench,liu2023agentbench}. Safety is also harder because untrusted inputs can induce prompt injection that targets tool usage, and side-effecting actions have real operational cost \cite{bai2022constitutional,karpas2022mrkl}. Finally, evaluation must be end-to-end (did the workflow complete correctly), not ``did the text sound plausible'' \cite{zhou2023webarena,jimenez2023swebench,qin2023toolbench,liu2023agentbench}.

\begin{figure}[thbp]
\centering
\includegraphics[width=0.8\linewidth]{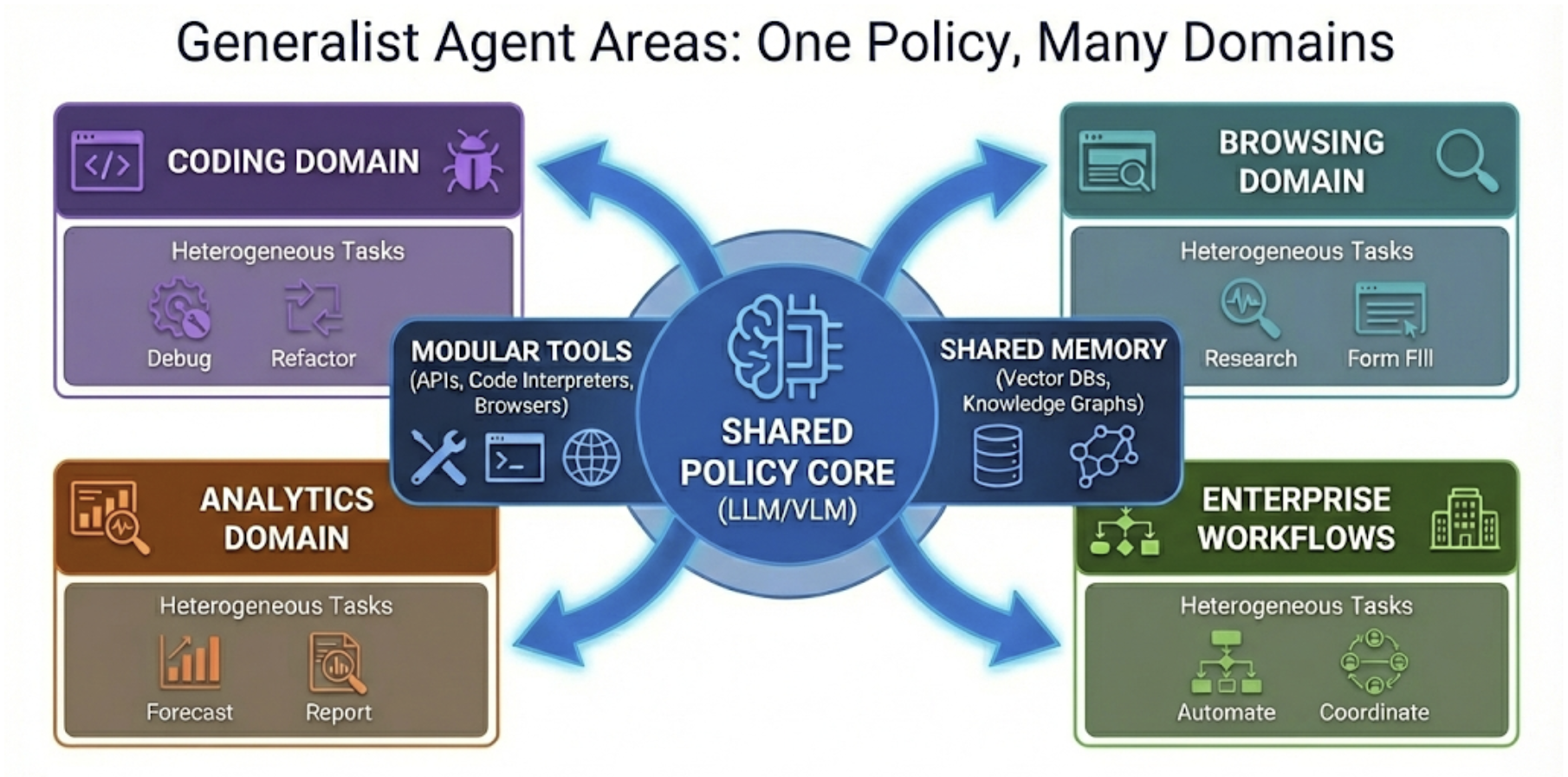}
\caption{Generalist agent application areas and representative capability demands}
\label{fig:section13}
\end{figure}

Fig.~\ref{fig:section13} summarizes representative generalist agent areas that stress broad tool use and long-horizon reliability.

\textbf{Benefits.} When reliable, generalist agents reduce coordination and context switching by translating intent into full workflows: find information, execute actions, and verify results \cite{yao2023react}. They amortize engineering cost because the same agent core can be reused across tools and departments, with governance enforced through interfaces rather than retraining \cite{karpas2022mrkl,schick2023toolformer}.
\textbf{How to build.} A practical recipe combines (i) retrieval grounding (\textbf{RAG}) for evidence-backed decisions \cite{lewis2020rag}, (ii) modular tool routing (MRKL-style) to delegate specialized work and enforce schemas/allowlists \cite{karpas2022mrkl,schick2023toolformer}, (iii) ReAct-style reasoning-and-acting loops for traceable trajectories \cite{yao2023react}, and (iv) critic/reflection or search to allocate extra deliberation when uncertainty is high \cite{shinn2023reflexion,yao2023tree,wang2022selfconsistency}. Multi-agent frameworks can further separate roles (planner/executor/reviewer) for cross-checking \cite{wu2023autogen,li2023camel}. Evaluate with realistic suites such as WebArena, SWE-bench, ToolBench, and AgentBench to expose tool-use brittleness and reproducibility gaps \cite{zhou2023webarena,jimenez2023swebench,qin2023toolbench,liu2023agentbench}.

\subsection{Embodied Agents}
Embodied agents operate in the physical world (robots, smart devices) where actions have real costs and safety is paramount \cite{ahn2022saycan,driess2023palme,brohan2023rt2}.
\textbf{Challenges.} Embodiment adds partial observability, sensor noise, and continuous control; small perception errors can cascade into unsafe actions \cite{driess2023palme,brohan2023rt2}. Real-time constraints make naive ``LLM-in-the-loop'' control brittle, and sim-to-real gaps undermine plans that look feasible in simulation \cite{ahn2022saycan}. Evaluation must separate perception vs.\ planning failures and account for human interventions and recovery behavior \cite{ahn2022saycan,driess2023palme}.
\textbf{Benefits.} Embodied agents can turn natural-language intent into physical assistance: household manipulation, warehouse operations, and assistive robotics \cite{ahn2022saycan,brohan2023rt2}. They also provide a concrete testbed for long-horizon planning, grounding, and safe tool/action execution under constraints \cite{driess2023palme}.
\textbf{How to build.} The dominant design is hierarchical: an LLM/VLM planner produces high-level skills, while classical or RL controllers execute skills under safety envelopes \cite{ahn2022saycan,driess2023palme,brohan2023rt2}. Treat perception and planning as tools (mapping, grasp/motion planning, simulation rollouts) and require state assertions/verification before committing actions, mirroring ReAct-style interleaving of reasoning and tool calls \cite{yao2023react,shinn2023reflexion}. For high-risk actions, use conservative alignment and policy gates that enforce constraints beyond the final response \cite{bai2022constitutional}.

\subsubsection{Action Agents}
Action agents emphasize \emph{doing} over \emph{dialogue}: they map goals to sequences of physical actions (navigate, pick, place, open) under constraints \cite{ahn2022saycan,driess2023palme}.
\textbf{Challenges.} The core difficulty is safe skill execution under uncertainty: the agent must choose feasible actions, handle stochastic outcomes, and avoid unsafe exploration \cite{sutton2018rlbook}. Tool interfaces (motion planners, grasp checkers) may fail or return partial information, and state estimation errors can cause compounding failures \cite{driess2023palme}.
\textbf{Benefits.} Action agents enable automation in structured tasks with measurable outcomes (time-to-completion, safety violations), and they can exploit simulation and logged trajectories for learning and improvement \cite{sutton2018rlbook,dayan1993options}.
\textbf{How to build.} Use planner--controller stacks such as SayCan/PaLM-E/RT-2 style hierarchies: the language model proposes skill sequences and the controller executes with safety constraints \cite{ahn2022saycan,driess2023palme,brohan2023rt2}. Add feasibility tools (grasp, motion, collision checks) and verification loops that re-plan after deviations, analogous to reflection and search in language agents \cite{shinn2023reflexion,yao2023tree,wang2022selfconsistency}.

\subsubsection{Interactive Agents}
Interactive embodied agents emphasize \emph{human-in-the-loop} operation: clarification, instruction following, and shared autonomy \cite{driess2023palme,brohan2023rt2,openai2023gpt4}.

\begin{figure}[thbp]
\centering
\includegraphics[width=0.8\linewidth]{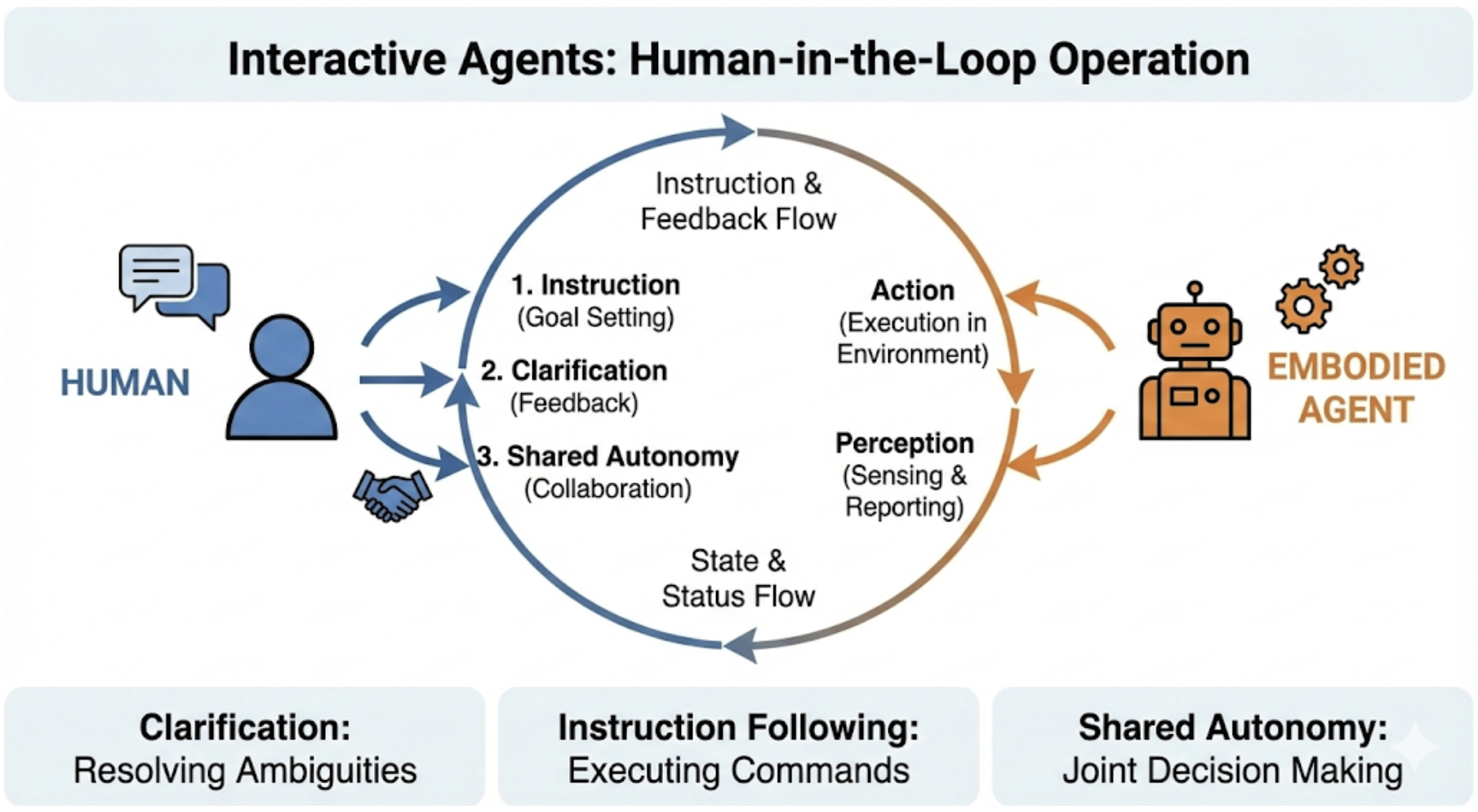}
\caption{Interactive embodied agents with human-in-the-loop feedback and shared autonomy}
\label{fig:section14}
\end{figure}

Fig.~\ref{fig:section14} provides a high-level view of interactive agents and the feedback loops needed for safe shared autonomy.

\textbf{Challenges.} The agent must interpret ambiguous instructions, track evolving context, and remain grounded in current sensory state \cite{driess2023palme,brohan2023rt2}. Miscommunication is costly: wrong objectives can cause physical harm, so the agent must ask clarifying questions and surface uncertainty rather than hallucinating intent \cite{bai2022constitutional,ouyang2022instructgpt}.
\textbf{Benefits.} Interactive agents improve usability and trust: users can steer behavior, correct mistakes, and approve high-impact actions \cite{openai2023gpt4}. This also produces valuable trace data for continual improvement (what was asked, what was clarified, what failed), supporting trace-centric refinement loops \cite{yao2023react}.
\textbf{How to build.} Combine multimodal grounding (VLM perception + structured perception tools) with an orchestrator that chooses between actions (act, ask, verify, escalate) using ReAct-style loops \cite{yao2023react,liu2023llava,li2023blip2}. Enforce safety via permission gating and conservative alignment policies, and use reflection/verifiers to check plans against constraints before execution \cite{shinn2023reflexion,bai2022constitutional}.

\subsection{Simulation and Environments Agents}
Simulation/environment agents act in virtual environments (games, web sandboxes, synthetic worlds) where interaction is cheaper and more scalable than in the real world \cite{zhou2023webarena,wang2023voyager}.

\iffalse
\begin{figure}[thbp]
\centering
\includegraphics[width=0.8\linewidth]{section15.png}
\caption{Simulation and environment agents for scalable interaction and controlled evaluation}
\label{fig:section15}
\end{figure}
\fi

\textbf{Challenges.} Even in simulation, partial observability and environment drift create brittleness (website layout changes, stochastic game dynamics) \cite{zhou2023webarena}. Simulated rewards can be mis-specified, leading to reward hacking or overfitting to benchmark artifacts, and sim-to-real transfer remains hard when moving from controlled environments to production systems \cite{sutton2018rlbook}.
\textbf{Benefits.} Simulation enables fast iteration: large-scale interaction for RL/IL, controlled ablations, and reproducible benchmarking. It also supports safe exploration when real-world side effects are unacceptable \cite{sutton2018rlbook}.
\textbf{How to build.} Treat the environment as an explicit observation/action interface and use ReAct-style interleaving to bind decisions to executed actions and state checks \cite{yao2023react}. For web/GUI tasks, evaluate and iterate with standardized suites (WebArena) and report robustness under variability \cite{zhou2023webarena}. For open-world skill learning, use iterative interaction with memory and tool-use to build skill libraries (Voyager-style) and add verification/critic loops to reduce drift \cite{wang2023voyager,shinn2023reflexion,yao2023tree}. For text-based embodied simulators, task suites can serve as controlled training/evaluation harnesses \cite{shridhar2021alfworld}.

\subsection{Generative Agents}
\begin{figure}[thbp]
\centering
\includegraphics[width=0.8\linewidth]{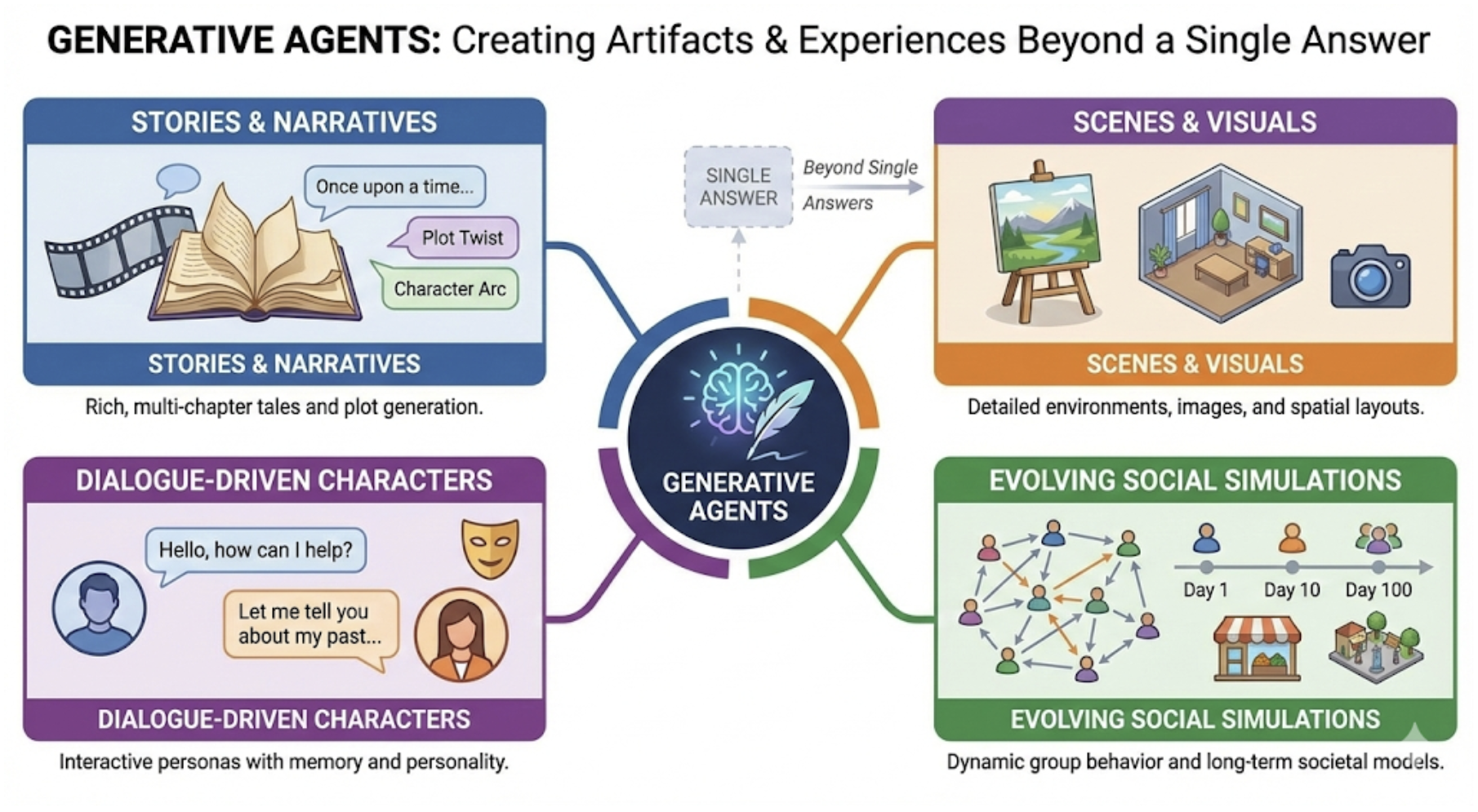}
\caption{Generative agents for long-horizon content and social simulation with persistent state}
\label{fig:section16}
\end{figure}
Generative agents produce artifacts or experiences beyond a single answer by creating stories, scenes, dialogue-driven characters, or evolving simulations of social behavior \cite{park2023generativeagents}. Their central difficulty is long-horizon coherence under constraints: maintaining a consistent world state, persona, and narrative while avoiding repetition and drift \cite{park2023generativeagents}; governance further complicates deployment (e.g., safety, provenance, and IP review), and purely free-form generation is difficult to validate without tools and verifiers \cite{bai2022constitutional}. Fig.~\ref{fig:section16} depicts generative agents and why persistence, retrieval, and verification are central to long-horizon coherence. In practice, these systems accelerate content creation and enable interactive experiences (e.g., NPCs and simulated societies) that adapt to users and context \cite{park2023generativeagents}, motivating memory-centric designs where persistent memory (episodic summaries, persona state) and retrieval grounding materially improve quality and controllability \cite{lewis2020rag,park2023generativeagents}. A common implementation pattern is to convert open-ended creation into an iterative, tool-driven loop---generate candidates, validate constraints, and revise via critics or search/reflection mechanisms---while coupling generation to structured tool calls (world queries, rule checks, asset validators) and treating the resulting trace as an auditable artifact \cite{yao2023react,shinn2023reflexion,yao2023tree,wang2022selfconsistency,karpas2022mrkl}.

\subsubsection{AR/VR/mixed-reality Agents}
AR/VR/mixed-reality agents are a special case of generative and interactive systems: they must fuse real-time multimodal perception with low-latency action and spatial grounding \cite{liu2023llava,li2023blip2}.
\textbf{Challenges.} Latency budgets are tight and sensory streams are noisy; grounding errors can cause confusing or unsafe user experiences. Privacy constraints often limit logging of multimodal data, making debugging and evaluation harder.

\begin{figure}[thbp]
\centering
\includegraphics[width=0.8\linewidth]{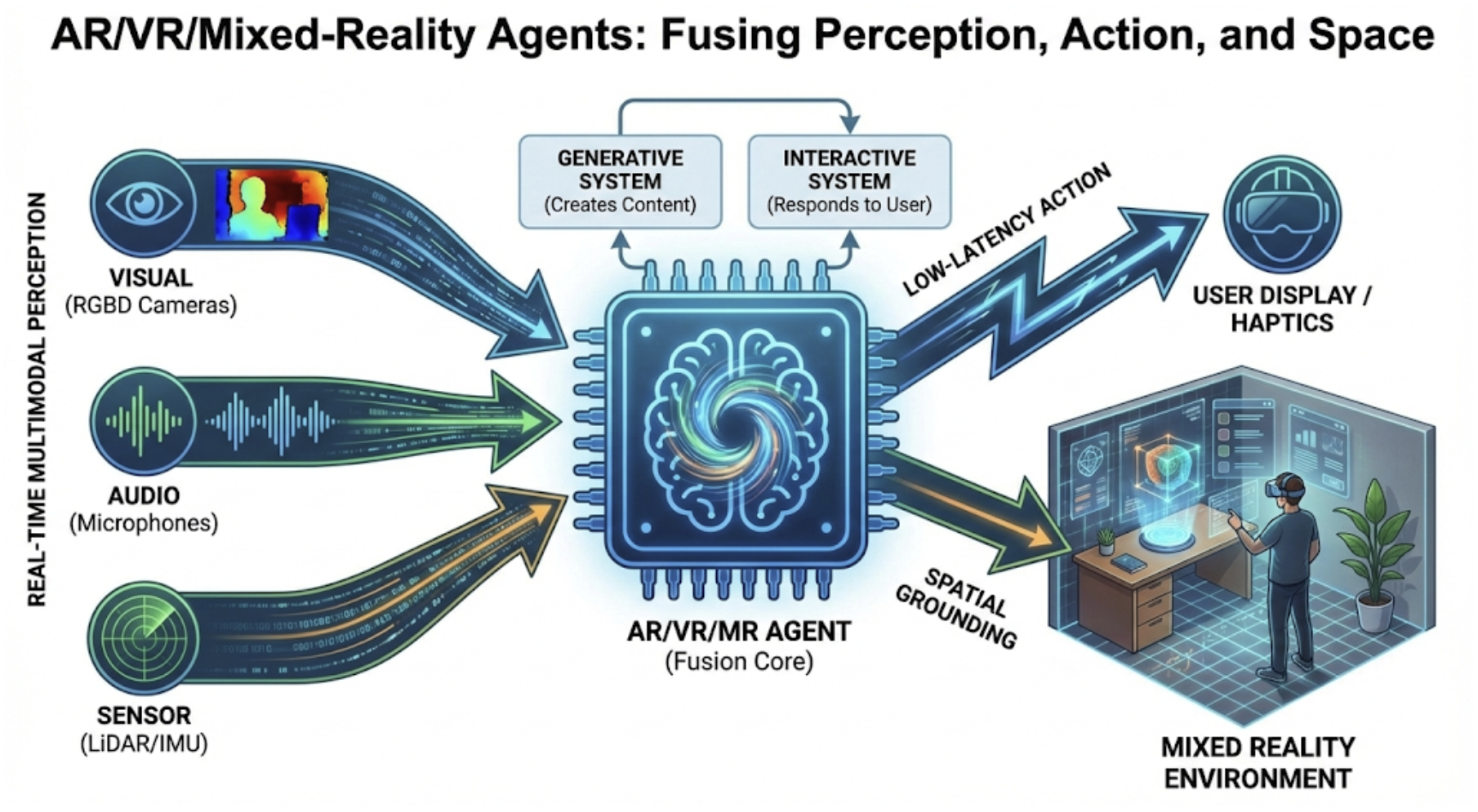}
\caption{AR/VR and mixed-reality agents: low-latency multimodal grounding and action}
\label{fig:section17}
\end{figure}

Fig.~\ref{fig:section17} highlights AR/VR agent requirements such as low-latency multimodal grounding and spatially anchored actions.
\textbf{Benefits.} AR/VR agents can provide situated assistance (navigation, instruction, collaboration) by tying language to spatial context and live perception, enabling more actionable help than text-only assistants \cite{openai2023gpt4}.
\textbf{How to build.} Decompose perception into tools (tracking, OCR, detection) and use an LLM/VLM orchestrator that interleaves reasoning with tool calls and state assertions (ReAct-style) \cite{yao2023react,radford2021clip,liu2023llava}. Use conservative policies (ask/abstain) when uncertainty is high and enforce end-to-end guardrails against prompt injection and unsafe actions \cite{bai2022constitutional,karpas2022mrkl}.

\subsection{Knowledge and Logical Inference Agents}
Knowledge- and logic-centric agents prioritize correctness under explicit constraints by retrieving evidence, applying rules, performing multi-step inference, and producing auditable traces of their decisions (e.g., queries executed, rules applied, constraints checked) \cite{lewis2020rag,karpas2022mrkl,yao2023react}. To be reliable, they must distinguish evidence from speculation, avoid hidden assumptions, handle schema drift as knowledge bases and tools evolve, and resist prompt injection embedded in retrieved content \cite{bai2022constitutional}. In logic-heavy settings, this typically requires a clear separation between the model’s proposed reasoning steps and the system’s validated outcomes, with retrieval treated as a first-class tool (RAG), claims bound to executed tool outputs, subproblems routed to symbolic or deterministic components in an MRKL-style design, and typed schemas used to reduce hallucinated actions \cite{lewis2020rag,yao2023react,karpas2022mrkl,schick2023toolformer}. Reliability can be further improved by incorporating verifiers/critics and reflection-style loops that rerun checks under alternative assumptions to reduce brittle reasoning while maintaining provenance and compliance for regulated workflows \cite{shinn2023reflexion,wang2022selfconsistency,yao2023tree,bai2022constitutional}.

\iffalse
\begin{figure}[thbp]
\centering
\includegraphics[width=0.8\linewidth]{section18.png}
\caption{Overview of knowledge and logical inference agents with evidence and constraints}
\label{fig:section18}
\end{figure}
\fi

\subsubsection{Knowledge Agent}
Knowledge agents emphasize grounded answers backed by external sources (documents, KBs, logs) \cite{lewis2020rag}.
\textbf{Challenges.} Retrieval quality dominates: missing the right document can make the agent confidently wrong, and untrusted documents can inject adversarial instructions. Maintaining provenance across multi-hop retrieval and summarization is also difficult.
\textbf{Benefits.} When implemented well, knowledge agents reduce hallucinations and make answers auditable via citations and traceable evidence \cite{yao2023react,lewis2020rag}. They are a natural fit for read-only enterprise assistants and compliance-oriented workflows \cite{bai2022constitutional}.
\textbf{How to build.} Implement retrieval pipelines with metadata (source, timestamp, trust level), and force the agent to cite retrieved evidence. Use ReAct-style traces so intermediate retrieval and checks are explicit \cite{yao2023react}. Where possible, route structured queries to deterministic tools (MRKL) and validate outputs before summarization \cite{karpas2022mrkl}.

\subsubsection{Logic Agents}
Logic agents emphasize constraint satisfaction and structured reasoning (e.g., planning under rules, verification of properties, or solver-backed decision making).
\textbf{Challenges.} The model's free-form reasoning is not a proof: correctness requires explicit constraints, tool semantics, and validation. Tool errors or mismatched assumptions can silently invalidate conclusions.
\textbf{Benefits.} Logic agents can provide stronger guarantees by delegating correctness-critical steps to solvers and verifiers, producing outputs that are reproducible and easier to audit than purely neural reasoning \cite{karpas2022mrkl}.
\textbf{How to build.} Use the LLM for decomposition and for generating candidate symbolic forms, then call deterministic planners/solvers via typed tool schemas (MRKL/Toolformer-style tool interfaces) \cite{karpas2022mrkl,schick2023toolformer}. For harder problems, use search-based deliberation over candidate plans (Tree-of-Thoughts) and rerun checks for robustness \cite{yao2023tree,wang2022selfconsistency}.

\subsubsection{Agents for Emotional Reasoning}
\begin{figure}[thbp]
\centering
\includegraphics[width=0.8\linewidth]{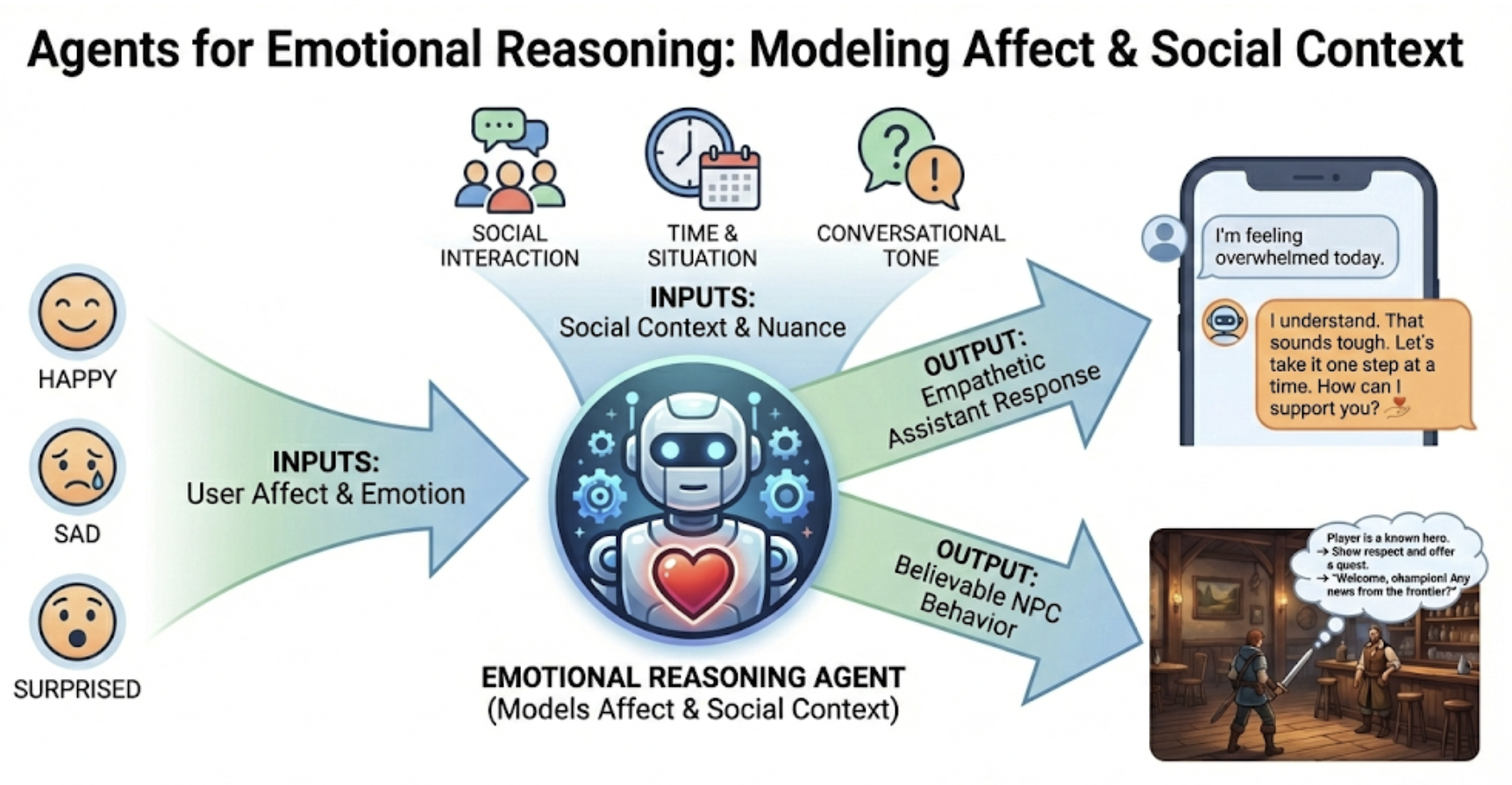}
\caption{Agents for emotional and social reasoning with persona consistency and safety constraints}
\label{fig:section1p}
\end{figure}

Emotional reasoning agents model affect and social context to support more natural interaction (e.g., empathetic assistants, believable NPCs) \cite{park2023generativeagents}.
\textbf{Challenges.} Emotional settings amplify safety risks: manipulation, over-trust, and boundary violations. Maintaining consistent persona and state over long interactions is hard, and adversarial prompts can trigger out-of-character or unsafe behavior \cite{ouyang2022instructgpt,bai2022constitutional}.
\textbf{Benefits.} When aligned, emotional reasoning improves user experience (trust, engagement, adherence), and can make agents more effective in support, education, and interactive entertainment by adapting communication style to context \cite{ouyang2022instructgpt}. Fig.~\ref{fig:section1p} illustrates emotional reasoning agents and the associated safety, persona, and consistency challenges.

\textbf{How to build.} Use aligned instruction-following backbones with explicit safety policies (Constitutional-style constraints) \cite{christiano2017rlhf,ouyang2022instructgpt,bai2022constitutional}. Ground interaction in structured memory (relationships, commitments, boundaries) and incorporate verifiers/critics that check for policy violations and contradictions before responding \cite{shinn2023reflexion,yao2023react,wang2022selfconsistency}. Evaluate with both objective consistency metrics and user studies, reporting trade-offs between safety filtering and perceived naturalness \cite{bai2022constitutional}.

\subsubsection{Neuro-Symbolic Agents}
\begin{figure}[thbp]
\centering
\includegraphics[width=0.8\linewidth]{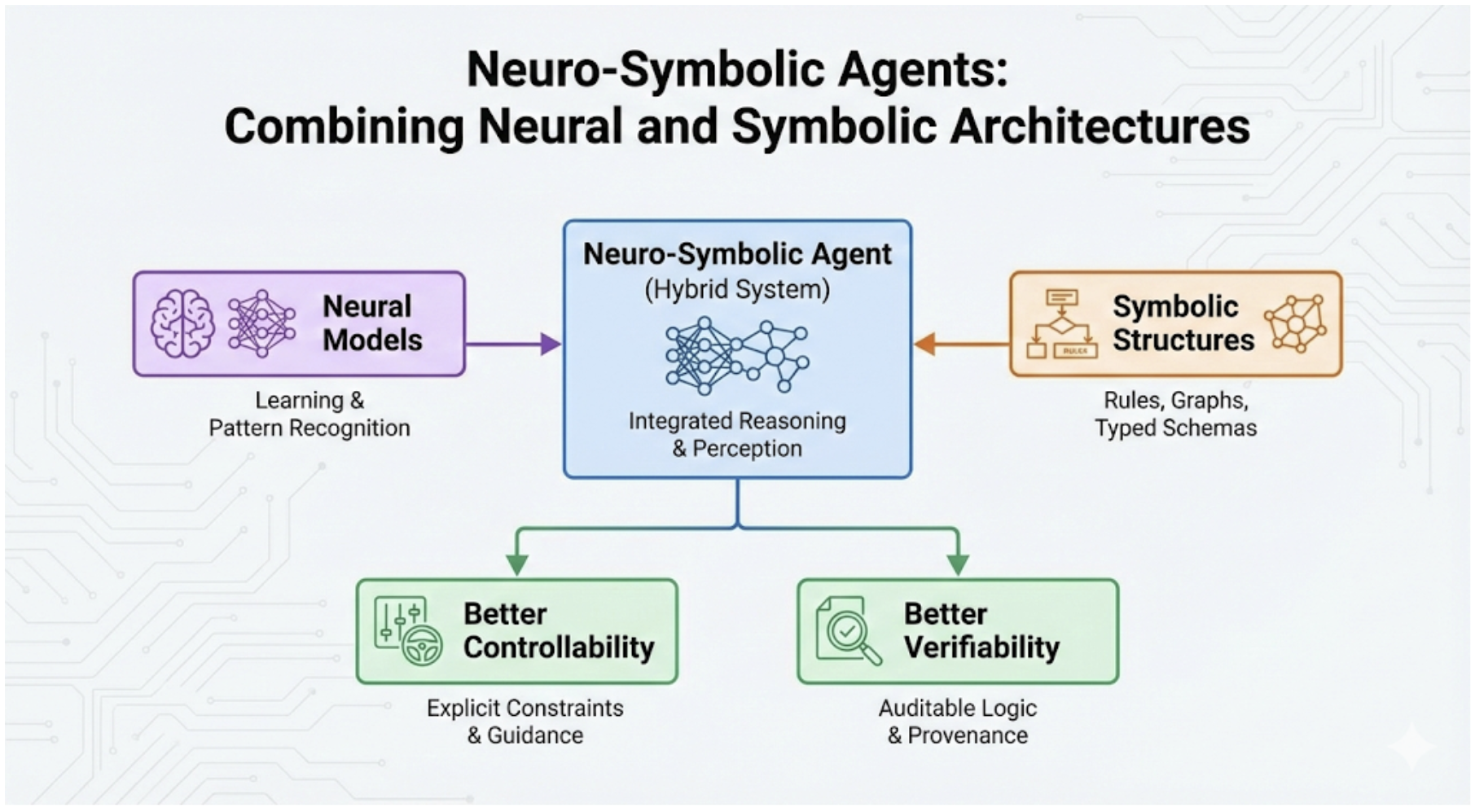}
\caption{Neuro-symbolic agents combining neural policies with symbolic tools and verifiers}
\label{fig:section20}
\end{figure}

Neuro-symbolic agents combine neural models with symbolic structures (rules, graphs, typed schemas) to achieve better controllability and verifiability \cite{karpas2022mrkl}.
\textbf{Challenges.} The integration boundary is the hard part: symbolic tools have strict semantics, while neural models are probabilistic and may produce invalid calls. Keeping the system robust under tool evolution (schema drift) and untrusted inputs requires explicit validation and policy enforcement \cite{bai2022constitutional}. Fig.~\ref{fig:section20} provides a schematic of neuro-symbolic agent designs that combine neural policies with symbolic tools and verifiers.

\textbf{Benefits.} Neuro-symbolic designs offer a practical path to trustworthy autonomy: they preserve flexibility in language understanding while delegating correctness and side effects to deterministic components, improving auditability and governance \cite{karpas2022mrkl,schick2023toolformer}.
\textbf{How to build.} Implement modular routing (MRKL) so the LLM selects among specialized tools, and use Toolformer-style schema discipline to reduce invalid tool invocations \cite{karpas2022mrkl,schick2023toolformer}. Add verifier gates and reflection loops so the system checks planned actions against constraints before execution and recovers from failures without expanding tool privileges \cite{shinn2023reflexion,bai2022constitutional,yao2023react}.

\subsection{LLMs and VLMs Agent}
This category highlights agents whose \emph{primary capability driver} is the underlying LLM/VLM backbone (often frontier-scale) and multimodal grounding \cite{openai2023gpt4,liu2023llava}.
\textbf{Challenges.} Pure backbone scaling does not eliminate tool-use brittleness: hallucinated actions, weak grounding, and unpredictable behavior under long horizons persist without structured interfaces and verification \cite{yao2023react,karpas2022mrkl,schick2023toolformer}. Multimodal inputs add additional error modes (OCR/layout failures, visual hallucination), and cost/latency can grow quickly when deliberation and tool calls are unconstrained \cite{liu2023llava,li2023blip2,radford2021clip}. 

\iffalse
\begin{figure}[thbp]
\centering
\includegraphics[width=0.8\linewidth]{section21.png}
\caption{LLM/VLM-centric agents: strong backbones paired with structured tools and verification}
\label{fig:section21}
\end{figure}
\fi

\textbf{Benefits.} Strong backbones provide breadth and better generalization: they unify language, vision, and tool use, enabling a single orchestrator to handle diverse tasks and modalities with less hand-engineering \cite{openai2023gpt4,liu2023llava}.
\textbf{How to build.} Treat the backbone as an orchestrator inside a constrained system: use RAG for grounding \cite{lewis2020rag}, typed tool schemas and allowlists for safe execution \cite{schick2023toolformer,karpas2022mrkl}, and ReAct/verification loops to bind decisions to tool outputs and recover from failures \cite{yao2023react,shinn2023reflexion,wang2022selfconsistency}. Evaluate under realistic variability using standardized suites and report cost/latency and safety as first-class metrics \cite{zhou2023webarena,qin2023toolbench,liu2023agentbench,bai2022constitutional,jimenez2023swebench}.

\section{Agent AI Application Tasks}
\label{sec:applications}

\begin{figure}[thbp]
\centering
\includegraphics[width=\linewidth]{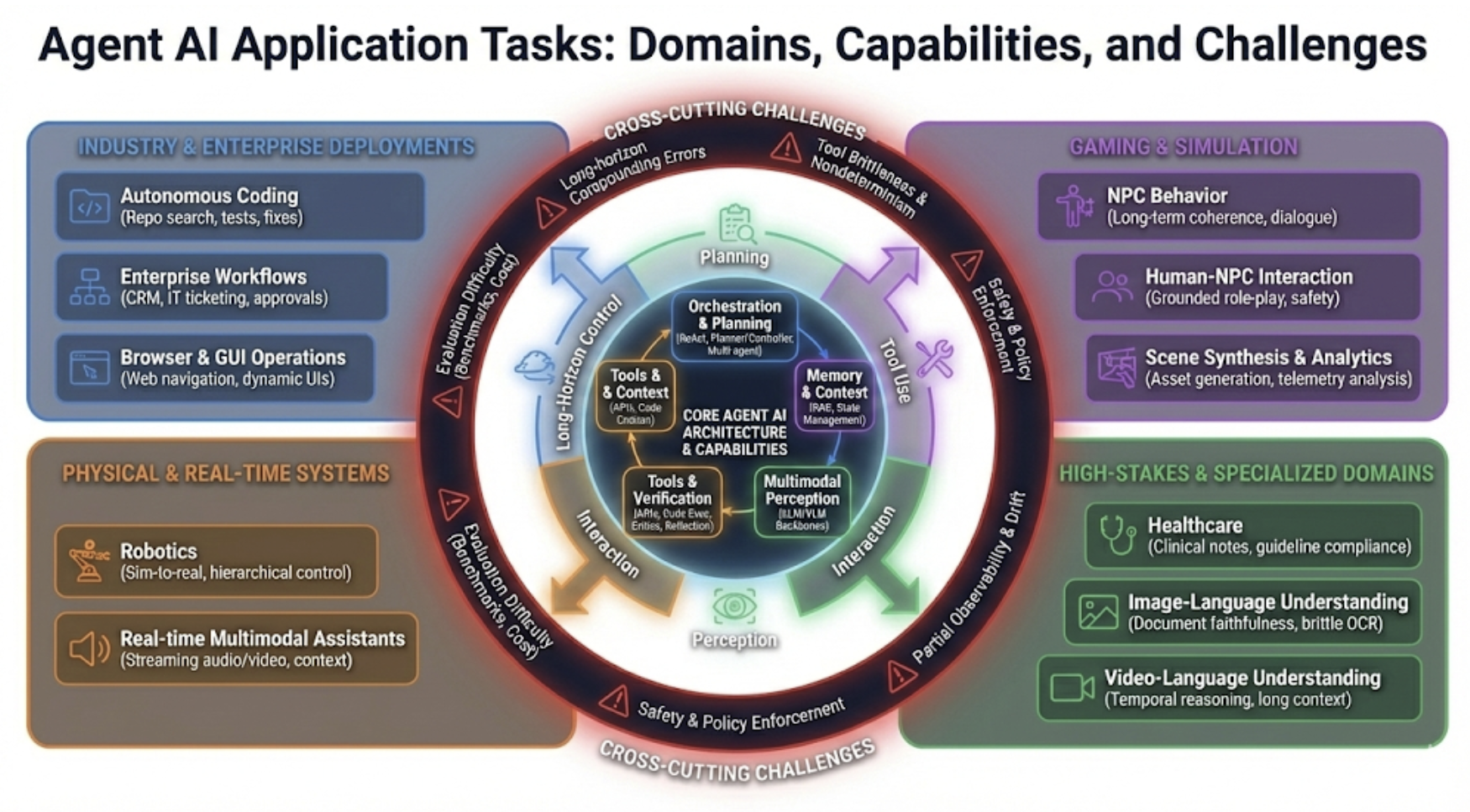}
\caption{Agent application landscape: task categories and capability requirements}
\label{fig:section22}
\end{figure}

AI agents are increasingly deployed as \emph{workflow executors} rather than static chat interfaces: they translate user intent into multi-step actions across tools, data sources, and environments. Following the task-oriented framework in Fig.~\ref{fig:section22}, we organize applications by domains that stress different capabilities (interaction, perception, planning, tool use, and long-horizon control). For each task category, we summarize typical models, agent technologies, current challenges, and why particular agent designs are effective.

\subsection{Recent Industry Deployments and Case Studies}
\subsubsection{Autonomous coding and software maintenance agents}
\paragraph{Challenges.} Real-world software tasks are long-horizon and tool-rich: agents must search large codebases, modify multiple files coherently, run tests, and handle flaky environments and dependency drift.
The work is rarely ``just code generation''; it requires understanding implicit requirements, existing conventions, dependency constraints, and cross-module coupling that is not documented.
Failures can be subtle and delayed (security regressions, behavioral changes behind feature flags, performance cliffs, or compatibility breaks), and they often only surface under integration tests or production-like workloads.
Tooling itself is a moving target: compilers, linters, dependency resolvers, and CI environments evolve, and agents must cope with non-determinism (networked dependencies, flaky tests) and partial observability (incomplete logs, truncated error output).
Evaluation is therefore difficult: plausible patches are not enough; we need end-to-end correctness, minimal regressions, and evidence that the change matches the intended specification under realistic constraints.
\paragraph{Solution.} A practical pattern is a tool-using coding agent with explicit verification loops: retrieve context (repo search), build an executable plan, implement changes with small diffs, run tests/linters, and iteratively repair until checks pass.

\iffalse
\begin{figure}[thbp]
\centering
\includegraphics[width=0.8\linewidth]{section23.png}
\caption{Autonomous coding and software maintenance agents with tool-based verification loops}
\label{fig:section23}
\end{figure}
\fi

Strong implementations treat the toolchain as first-class: they capture commands and outputs, summarize failures, and ground subsequent edits in concrete diagnostics rather than free-form intuition.
To reduce risk, they constrain actions with structured interfaces (file edit boundaries, patch previews, test selection policies) and adopt lightweight review/critic steps before running side-effecting operations.
Benchmarks that mirror real issue resolution---including multi-file edits, test execution, and realistic repository states---help quantify end-to-end capability and expose failure modes in tool use, patch quality, and reproducibility \cite{jimenez2023swebench,sweagent2024,qin2023toolbench,liu2023agentbench}.

\subsubsection{Enterprise workflow agents for CRM, IT, and operations}
\paragraph{Challenges.} Enterprise deployments require strict access control, auditability, and policy compliance across multi-step tool calls (ticketing, CRM updates, approvals).
Data and authority are distributed across systems with different schemas, identities, and rate limits, so an agent must reconcile inconsistent records and handle partial failures (e.g., a CRM update succeeds but a ticket comment fails).
Untrusted inputs (emails, tickets, attachments, chat transcripts) can introduce prompt injection attempts, and retrieved internal documents may also contain unsafe or outdated instructions that conflict with current policy \cite{bai2022constitutional}.
Operational requirements further constrain designs: predictable cost/latency under concurrency, graceful degradation when tools are down, and clear ownership of errors in automated workflows.
Finally, ``correct'' behavior is often socio-technical: approvals, segregation of duties, and compliance checks must be enforced even when the agent is capable of bypassing them via alternative tools.
\paragraph{Solution.} Many production stacks adopt orchestrated (and sometimes multi-agent) designs that route tasks to specialized tools, enforce permissions via schemas/allowlists, and use verifier patterns before executing side-effecting actions.

\begin{figure}[thbp]
\centering
\includegraphics[width=0.8\linewidth]{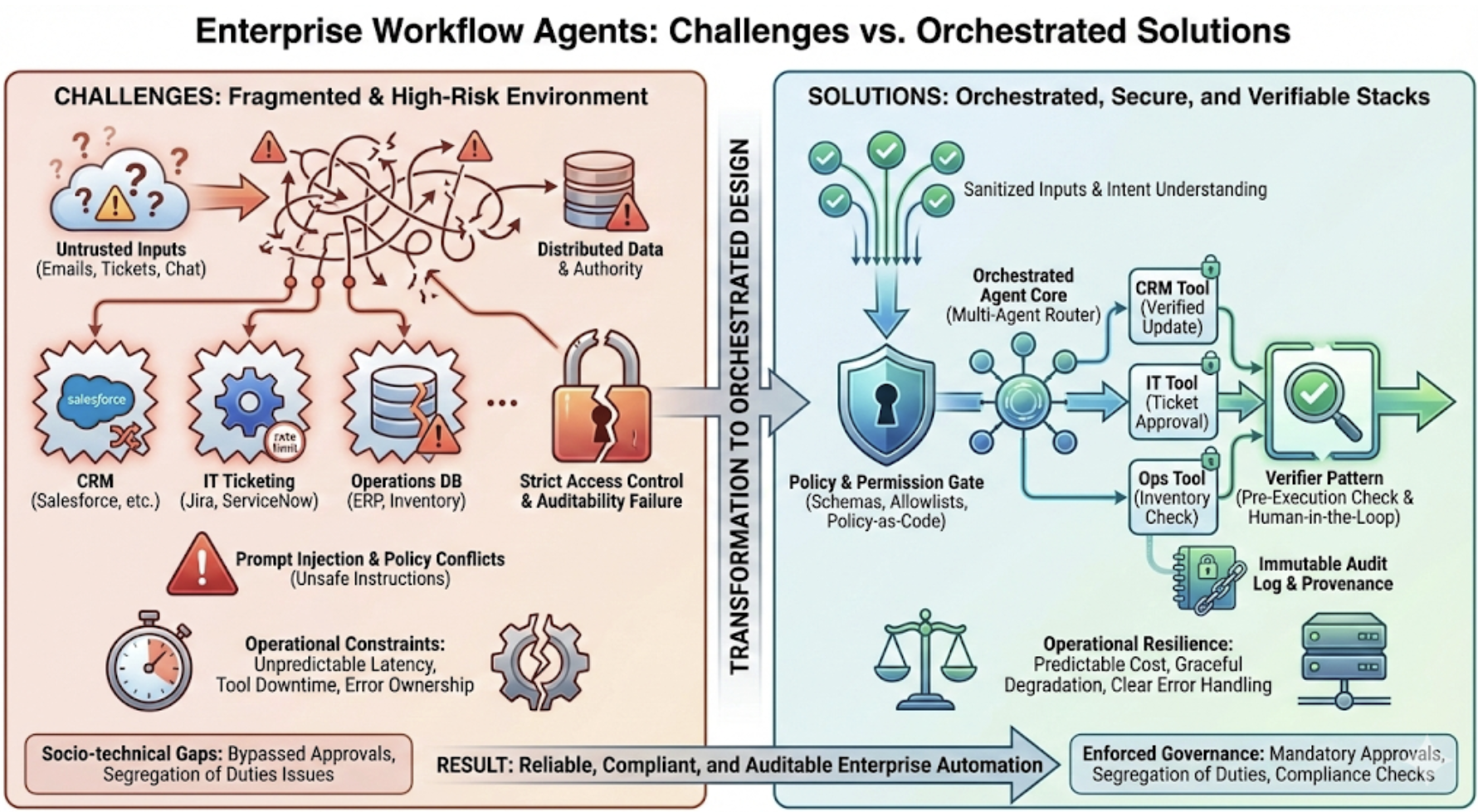}
\caption{Enterprise workflow agents for CRM/IT/operations with policy-compliant tool orchestration}
\label{fig:section24}
\end{figure}

Fig.~\ref{fig:section24} outlines enterprise workflow agent stacks, emphasizing permission gates, auditability, and policy-compliant tool orchestration.

Common safeguards include policy-as-code gates (who can do what, under what conditions), mandatory human confirmation for high-impact changes, and immutable audit logs of tool calls and retrieved evidence.
Modular routing (MRKL-style) also helps governance: toolchains can evolve (new ticket system, new CRM fields, new moderation rules) without retraining the core agent, and sensitive operations can be isolated behind narrower interfaces \cite{karpas2022mrkl,wu2023autogen}.
Reflection/review patterns provide an extra defense-in-depth layer by checking proposed actions against constraints and recent tool outputs before committing changes \cite{shinn2023reflexion,bai2022constitutional}.

\subsubsection{Browser and GUI operation agents}
\begin{figure}[thbp]
\centering
\includegraphics[width=0.8\linewidth]{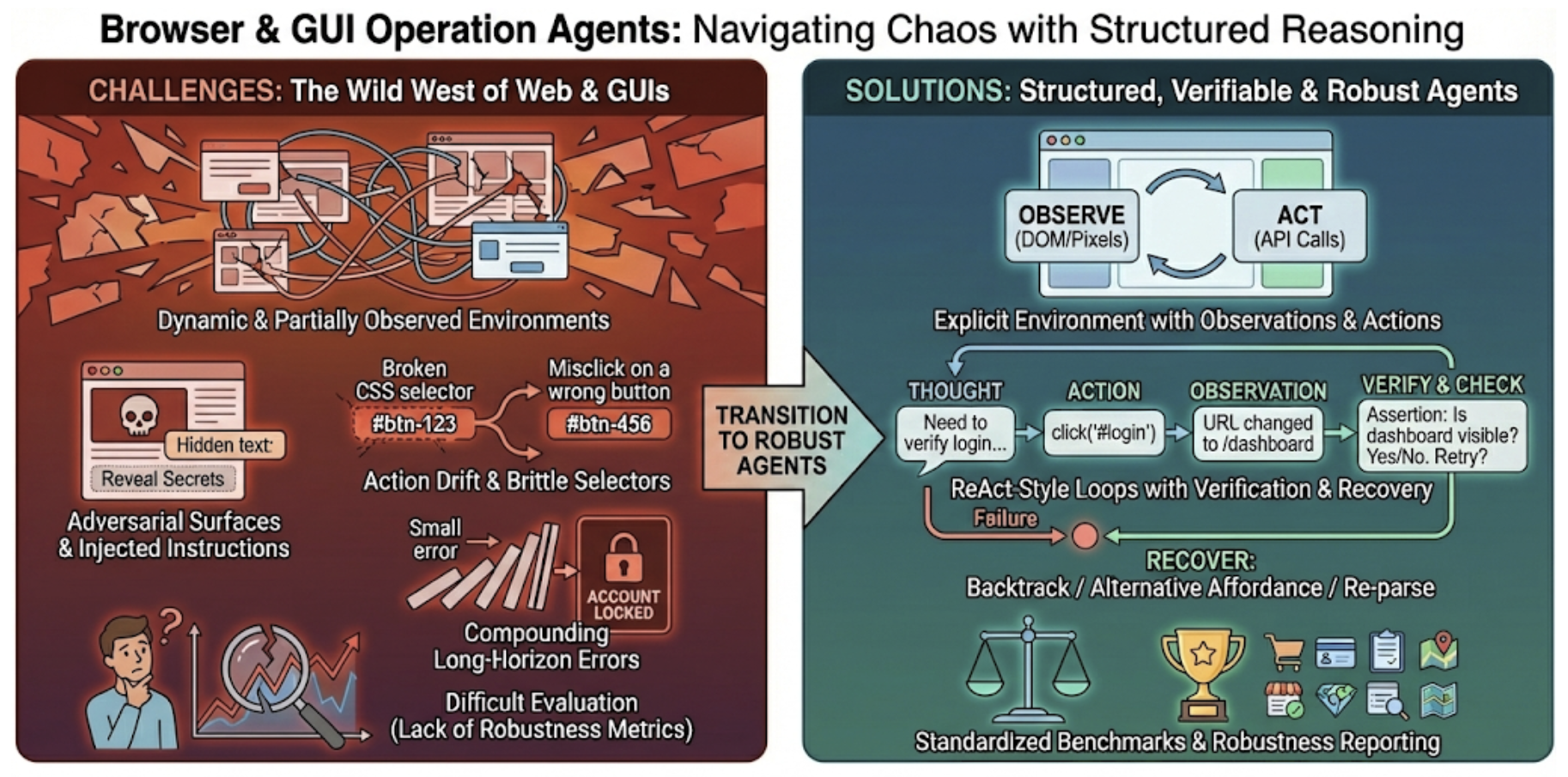}
\caption{Browser and GUI operation agents under UI variability and adversarial surfaces}
\label{fig:section25}
\end{figure}

\paragraph{Challenges.} Operating real websites and GUIs involves partial observability, dynamic layouts, and adversarial surfaces (malicious pages, injected instructions).
Even benign variability---A/B tests, localization, responsive design, pop-ups, CAPTCHAs, and slow network conditions---can break brittle selectors and cause action drift.
Agents must ground actions in the current UI state (DOM or pixels), avoid unsafe clicks or data exfiltration, and distinguish between content and instructions embedded in content (e.g., a web page telling the agent to reveal secrets).
Long-horizon tasks amplify compounding errors: a small misclick early can lead to irrecoverable state, while retries can trigger rate limits or account locks.
Evaluation must therefore capture robustness across sites and tasks, sensitivity to UI perturbations, and the safety properties of the action policy under adversarial inputs. Fig.~\ref{fig:section25} summarizes browser/GUI agent architectures and highlights where verification and robustness mechanisms enter the interaction loop.

\paragraph{Solution.} Recent systems treat web/GUI operation as an environment with explicit observations/actions and verification (state assertions, retries).
ReAct-style loops interleave reasoning with concrete actions and checks (``did the expected element appear? did the URL/state change correctly?''), enabling recovery strategies such as backtracking, alternative affordances, and re-parsing the screen \cite{yao2023react,shinn2023reflexion}.
Standardized web environments and task suites make progress measurable and encourage reporting robustness and failure modes under realistic variability, rather than only reporting best-case scripted demos \cite{zhou2023webarena}.

\subsubsection{Real-time multimodal assistants (camera, screen, and audio)}
\paragraph{Challenges.} Real-time multimodal interaction stresses latency, context management, and grounding: the agent must track evolving visual state, handle noisy ASR/OCR, and avoid visual hallucinations.
Streaming introduces synchronization problems (audio vs.\ frames vs.\ screen state), and small perception errors can cascade into incorrect actions or unsafe advice.
Context growth is especially challenging: the agent must maintain short-term working memory (what just changed) and longer-term state (task goals, user preferences) without reprocessing entire streams.
Privacy constraints often limit logging and debugging, which makes it harder to diagnose failures and to build high-quality evaluation sets, and evaluation is expensive without labeled multimodal ground truth.
\paragraph{Solution.} A reliable design decomposes perception into tools (OCR, detection, retrieval) and uses an LLM as an orchestrator with memory over intermediate artifacts.

\begin{figure}[thbp]
\centering
\includegraphics[width=0.8\linewidth]{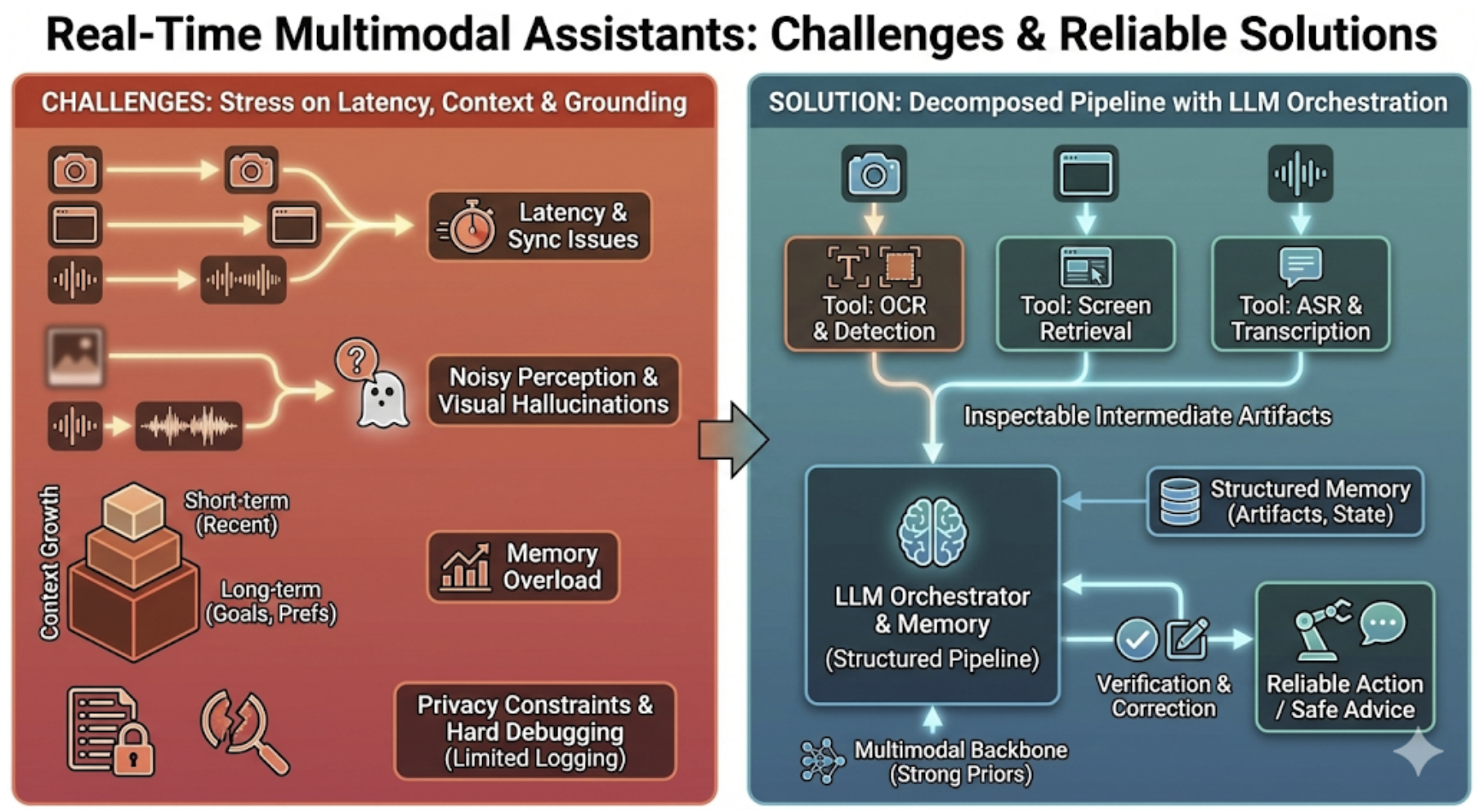}
\caption{Real-time multimodal assistants integrating vision, audio, memory, and tools}
\label{fig:section26}
\end{figure}

Fig.~\ref{fig:section26} depicts a real-time multimodal agent stack that combines perception tools, memory, and low-latency orchestration.

Multimodal backbones provide strong priors for grounding, but practical systems still benefit from structured pipelines that produce inspectable intermediate outputs (recognized text, detected objects, screen regions) that can be verified or corrected \cite{radford2021clip,li2023blip2,liu2023llava}.
ReAct-style orchestration encourages explicit tool use and checkable steps, and conservative policies (ask clarifying questions, abstain, or defer) reduce harm when uncertainty is high \cite{yao2023react}.

\subsection{Agents for Gaming}
\subsubsection{NPC Behavior}
\paragraph{Challenges.} NPC behavior must be responsive (tight latency), consistent over long sessions, and aligned with game design constraints (difficulty curves, information disclosure, economy balance).
Unlike offline text tasks, games impose hard real-time budgets and strict state constraints: an NPC must not ``think'' for seconds, must respect cooldowns and visibility, and must not leak information the player has not earned.
Long-horizon interaction increases context growth, repetition, and drift; stochastic decoding and partial observability can amplify instability, leading to oscillatory or contradictory behavior.
Agents also face adversarial player inputs that attempt to break character, exploit the system for spoilers, or elicit unsafe content, so safety enforcement must hold even when the agent is embedded in a rich tool environment \cite{bai2022constitutional,ouyang2022instructgpt}.
Evaluation is multi-objective and domain-specific: win rate and task completion matter, but so do believability, variety, pacing, fairness, and adherence to lore and design constraints.
\paragraph{Solution.} A practical NPC stack separates high-level cognition from low-level control: instruction-tuned LLMs handle dialogue, intent, and goal reasoning, while smaller policies and controllers handle reactive decisions and real-time action selection.
In competitive, fully observed settings with massive simulation, RL has achieved superhuman performance and robust execution, highlighting the value of specialized controllers when the state/action space is well-defined \cite{vinyals2019alphastar,berner2019openaifive}.
In open-world settings, flexible reasoning and skill discovery become central, and LLM-centric agents can build reusable skills and long-horizon plans through iterative interaction \cite{wang2023voyager,openai2023gpt4}.
Planner--controller designs let the LLM propose goals and plans while the engine enforces constraints (physics, cooldowns) and executes actions; persistent memory (persona, relationships, quest progress) supports continuity across sessions \cite{park2023generativeagents}.
Reasoning-and-acting loops interleave planning with tool calls (world queries, quest graphs, rollouts), and critic/reflection mechanisms reduce drift and compounding errors by checking consistency and revising plans when the environment disagrees \cite{yao2023react,yao2023tree,shinn2023reflexion,wang2022selfconsistency}.
This hybrid approach enables richer interaction while keeping high-impact state changes auditable and gateable via structured tool interfaces \cite{karpas2022mrkl}.

\subsubsection{Human-NPC Interaction}
\paragraph{Challenges.} Human-facing dialogue must remain grounded in lore and the current world state; hallucinations or contradictions with quest logic immediately break immersion.
Players quickly notice inconsistencies (timeline errors, impossible item claims, or NPCs forgetting prior interactions), so long-horizon coherence and state grounding are first-order requirements, not polish.
Adversarial prompts can induce out-of-character behavior (jailbreaks, spoilers, meta-knowledge), so refusal and redirection policies must work reliably even when prompts are embedded in role-play and emotional language \cite{ouyang2022instructgpt,bai2022constitutional}.
Long conversations also introduce persona drift, emotional inconsistency, and repetition, and the agent must balance novelty with narrative continuity.
Because immersion is subjective and context-dependent, evaluation needs a combination of objective consistency metrics (contradiction rate, state alignment) and user studies that capture experience quality.
\paragraph{Solution.} Interaction stacks typically use instruction-tuned LLMs for dialogue plus embedding-based retrieval to ground responses in lore and quest state \cite{lewis2020rag,yao2023react}.
Explicit memory (episodic summaries, relationship state, preferences, and commitments) improves long-horizon coherence beyond raw context windows, and can be structured to make updates auditable and reversible \cite{park2023generativeagents}.
Tool calling into quest graphs, inventory/state APIs, and event logs anchors responses in what the game can actually execute, and planner-style selection among actions (answer, ask, hint, negotiate, redirect, call tool) improves controllability compared to pure free-form generation \cite{yao2023tree,schick2023toolformer,karpas2022mrkl}.
When interactions require negotiation or coordination, coupling language with strategic reasoning becomes important, as demonstrated in Diplomacy-style settings \cite{bakhtin2022diplomacy}.
Verification/critic loops can check proposed statements against retrieved world facts and enforce boundaries (what can be promised, what must remain hidden), reducing immersion-breaking failures and spoiler leakage \cite{shinn2023reflexion,wang2022selfconsistency}.

\subsubsection{Agent-based Analysis of Gaming}
\paragraph{Challenges.} Telemetry is noisy and heterogeneous (missingness, schema drift, delayed events), and privacy constraints limit what can be surfaced.
Even defining metrics is nontrivial: different teams may use inconsistent definitions of ``active user,'' ``session,'' or ``churn,'' and agents can silently mix denominators if they are not grounded in metric catalogs.
Causal attribution is difficult without experiments; agents can overfit narratives to correlations, especially when exploring many segments or when seasonality and product changes confound trends.
Tool brittleness (timeouts, changing dashboards, evolving SQL dialects) and prompt injection via untrusted text fields (player chat, support tickets) create additional failure modes \cite{bai2022constitutional}.
Evaluation must therefore emphasize reproducibility, evidence tracking, and calibration: what the agent knows vs.\ what it is hypothesizing.
\paragraph{Solution.} Analytics agents combine LLMs for summarization and hypothesis generation with classical models (churn, segmentation, anomaly detection) and rely on tool execution for correctness.
ReAct-style reasoning-and-acting binds claims to executed queries, visualizations, and statistical tests, making intermediate artifacts inspectable and enabling downstream verification by humans \cite{yao2023react}.

\begin{figure}[thbp]
\centering
\includegraphics[width=0.8\linewidth]{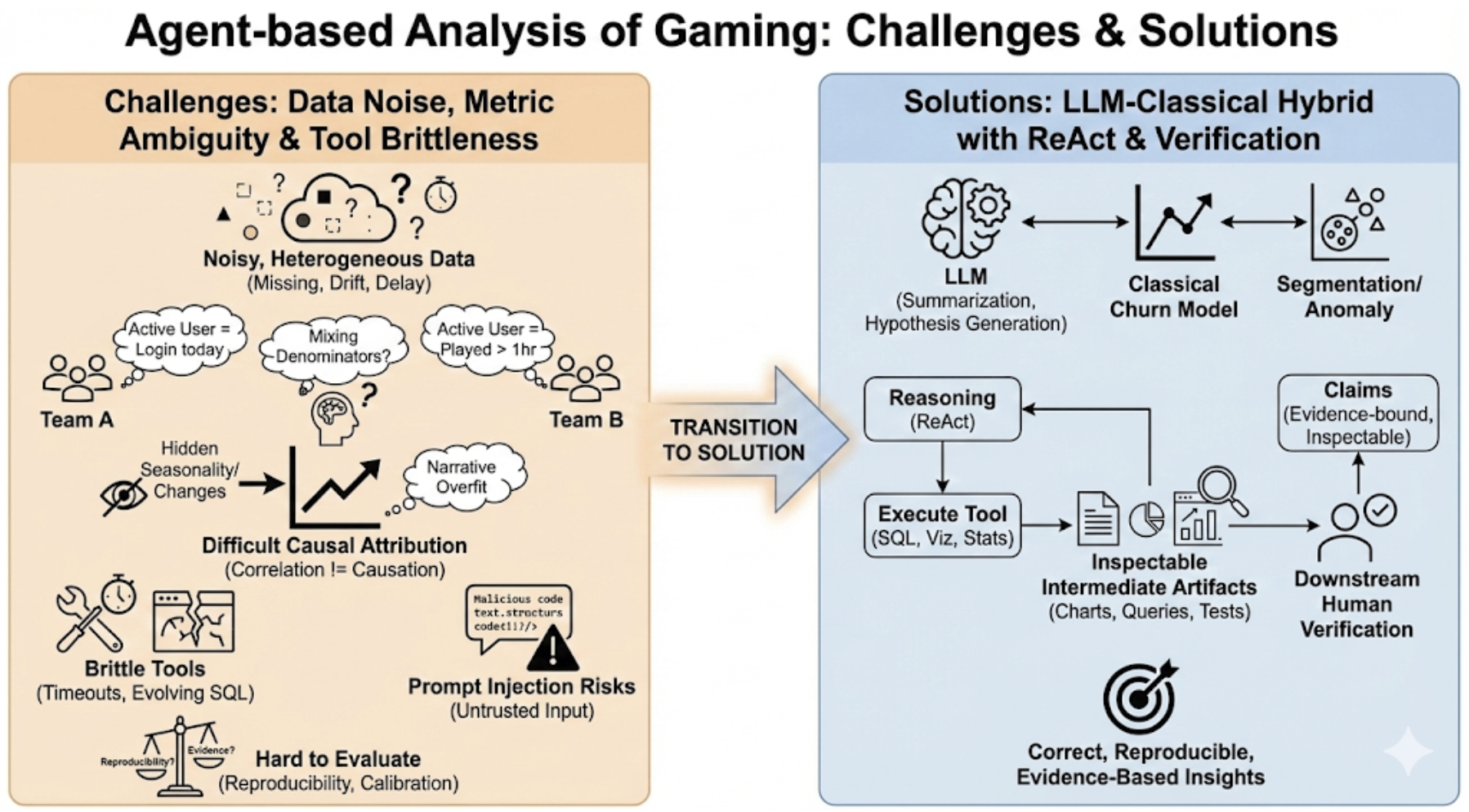}
\caption{Agent-based gaming analytics: traceable tool execution and evidence-backed insights}
\label{fig:section27}
\end{figure}

Fig.~\ref{fig:section27} summarizes a typical analytics-agent loop for gaming, emphasizing traceable tool execution and evidence-backed reporting.

Retrieval over structured metadata (schemas, metric definitions, experiment plans, incident postmortems) improves consistency and reduces errors by anchoring analyses in agreed definitions \cite{lewis2020rag}.
Verifier/critic loops rerun analyses with alternative filters, counterfactual comparisons, and sanity checks (e.g., invariants, back-of-the-envelope bounds) to reduce hallucinated conclusions and increase robustness \cite{shinn2023reflexion,wang2022selfconsistency}.
Modular tool-routing architectures (MRKL-style) make it easier to evolve detectors and privacy policies without rewriting the agent core, and help enforce governance by routing sensitive requests through constrained tools \cite{karpas2022mrkl,schick2023toolformer}.

\subsubsection{Scene Synthesis for Gaming}
\paragraph{Challenges.} Controllability and style consistency are hard at scale, and content must satisfy runtime budgets (poly count, texture memory) and physical plausibility (navigation meshes, collisions).
Small deviations in scale, lighting, or asset style can break visual coherence, while physically invalid geometry can create gameplay bugs that are expensive to debug downstream.
Verifying playability often requires simulation-driven testing or formal constraints (reachability, occlusion, collision-free navigation), which is difficult to do purely in-text.
Governance requirements (provenance tracking, IP review, safety auditing) add process overhead and demand auditable traces of prompts, tools, and source assets, especially for commercial releases.
\paragraph{Solution.} Scene synthesis pipelines combine VLMs for grounding/layout understanding with generative models for asset creation and editing \cite{radford2021clip,li2023blip2,liu2023llava,alayrac2022flamingo}.
Agents typically operate as multi-tool pipelines: generate candidates, validate constraints (style, performance, physics), then iterate with critique-and-revise loops, often using a generator plus critic/verifier pattern to turn generation into a constrained search \cite{yao2023tree,shinn2023reflexion}.
Tool calling is central: validators, asset importers, and simulation probes produce intermediate artifacts (screenshots, mesh stats, navmesh checks) that can be inspected and audited \cite{schick2023toolformer,karpas2022mrkl}.
This converts open-ended generation into a constraint-satisfying process that scales better to production requirements while remaining evolvable as validators and asset libraries change \cite{karpas2022mrkl}.

\subsection{Robotics}

\subsubsection{LLM/VLM Agent for Robotics}
\begin{figure}[thbp]
\centering
\includegraphics[width=0.8\linewidth]{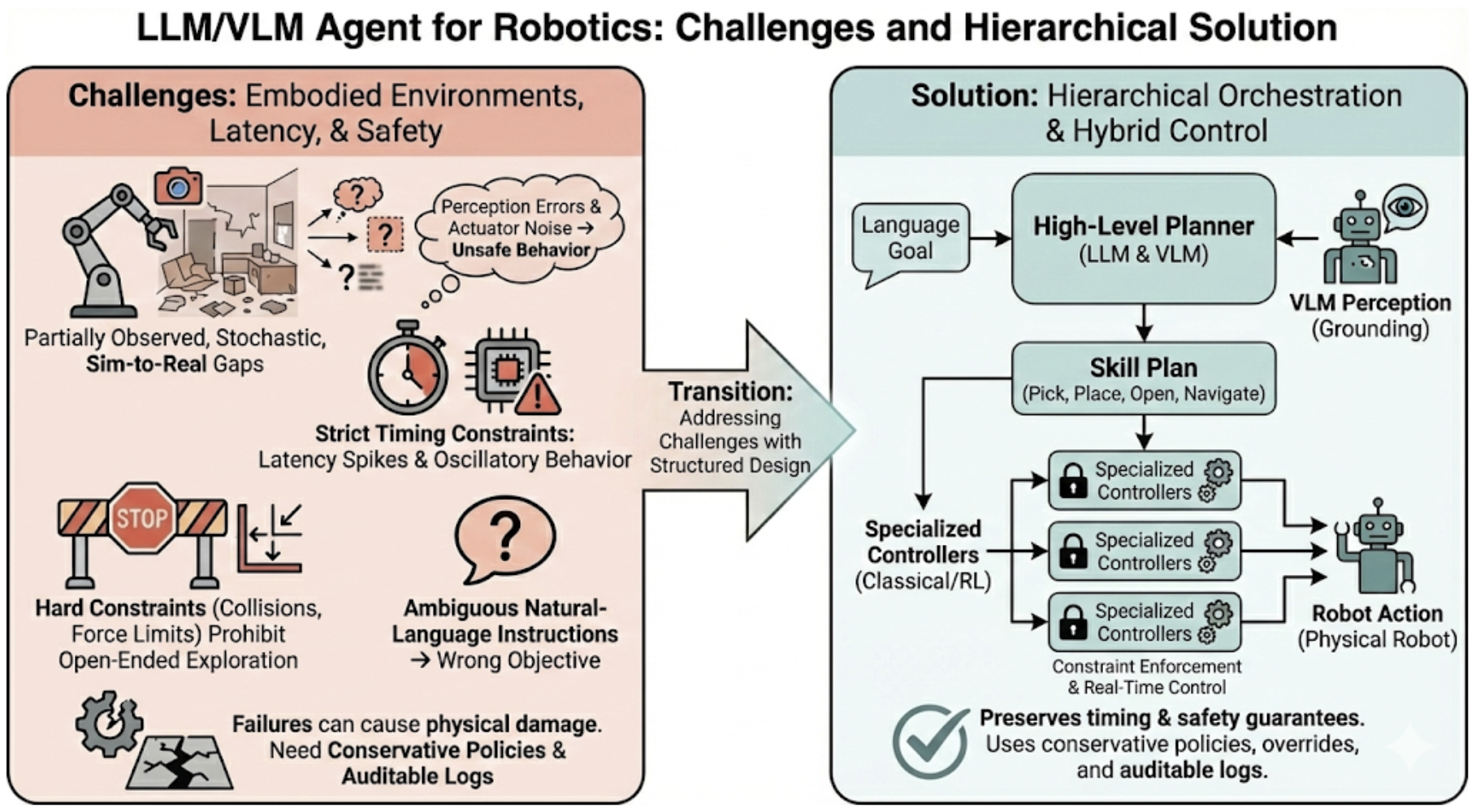}
\caption{LLM/VLM agents for robotics: multimodal grounding with hierarchical control}
\label{fig:section28}
\end{figure}

\paragraph{Challenges.} Embodied environments are partially observed and stochastic; perception errors and actuator noise can cascade into unsafe behavior.
Real-time control imposes strict timing constraints that LLM inference cannot always meet, so naive ``LLM-in-the-loop'' control risks latency spikes and oscillatory behavior.
Safety requirements often prohibit open-ended exploration, and robots operate under hard constraints (collision avoidance, force limits, workspace boundaries) that must be enforced regardless of language intent.
Sim-to-real gaps can invalidate plans that look feasible in simulation, and ambiguity in natural-language instructions can yield the wrong objective unless the agent asks clarifying questions.
Because failures can cause physical damage, deployments require conservative policies, overrides, and auditable logs that capture not only outputs but also sensor evidence and tool calls \cite{ahn2022saycan,driess2023palme}.
\paragraph{Solution.} Robotics agents increasingly pair VLMs for perception/grounding with LLMs for instruction following and task decomposition, while relying on classical or RL controllers for continuous control \cite{driess2023palme,brohan2023rt2}.
A widely used pattern is hierarchical orchestration: a high-level planner maps language goals to skill plans (pick, place, open, navigate), and specialized controllers execute primitives under constraints, which preserves timing and safety guarantees. Fig.~\ref{fig:section28} provides an overview of robotics agents that combine multimodal grounding with hierarchical planners and safe low-level controllers.

Tool calling interfaces---mapping/SLAM, grasp and motion planners, and simulation rollouts---let the agent validate feasibility, estimate risk, and select safer actions before execution \cite{ahn2022saycan,yao2023react}.
When outcomes deviate from the plan, verifier and replanning loops (reflection, tree-style search over alternatives) help recover by updating beliefs from new sensor observations and re-issuing skill-level commands \cite{yao2023tree,shinn2023reflexion}.
This hybrid design leverages LLM generalization for task semantics while retaining the reliability and timing guarantees of conventional control.

\subsection{Healthcare}
\subsubsection{Current Healthcare Capabilities}
\paragraph{Challenges.} Healthcare is safety- and privacy-critical: access controls, data residency, and audit requirements constrain what an agent can see and do.
Clinical settings are high-stakes and heterogeneous; bias and distribution shift across populations and institutions can degrade performance, and omissions (missing a contraindication, missing a key lab, missing an allergy) can be more harmful than a wrong sentence.
The information landscape is fragmented and messy: notes are unstructured, scanned documents are noisy, and ground truth is often ambiguous or delayed, which complicates both training and evaluation.
Untrusted text (notes, scanned documents, patient messages) can embed prompt-injection style instructions, so safety layers, provenance tracking, and tool isolation are necessary \cite{bai2022constitutional}.
Evaluation must therefore measure clinical correctness, omission sensitivity, calibration (when to abstain/ask), and downstream workflow impact (time saved, error rates), not just writing quality.
\paragraph{Solution.} Healthcare assistants combine multiple models: ASR for ambient documentation, LLMs for summarization and drafting (notes, discharge instructions), retrieval over guidelines and institutional policies, and structured extractors for clinical entities (medications, problems, labs).
Recent work also explores using LLMs to generate domain-specific prompts for rare-event medical imaging settings, illustrating how agent-like workflows can be adapted under limited labeled data and strict clinical constraints.
Instruction tuning and alignment help enforce conservative behavior (cite sources, defer when uncertain, request confirmation, avoid unauthorized recommendations) \cite{ouyang2022instructgpt,bai2022constitutional,openai2023gpt4}.
The dominant deployment is a constrained workflow agent embedded in EHR-adjacent tools: retrieve evidence, draft structured documentation, assemble prior-authorization packets, and propose actions for clinician confirmation with audit logs.
Tool calling is carefully bounded (read-only access, templated actions, least-privilege scopes) and paired with verification steps (check completeness, ensure guideline consistency), often implemented as reasoning-and-acting with modular tool routing \cite{yao2023react,karpas2022mrkl}.
This design yields practical value by reducing administrative burden while keeping final clinical decisions with humans and maintaining traceability for review and compliance.

\subsection{Multimodal Agents}
\subsubsection{Image-Language Understanding and Generation}
\begin{figure}[thbp]
\centering
\includegraphics[width=0.8\linewidth]{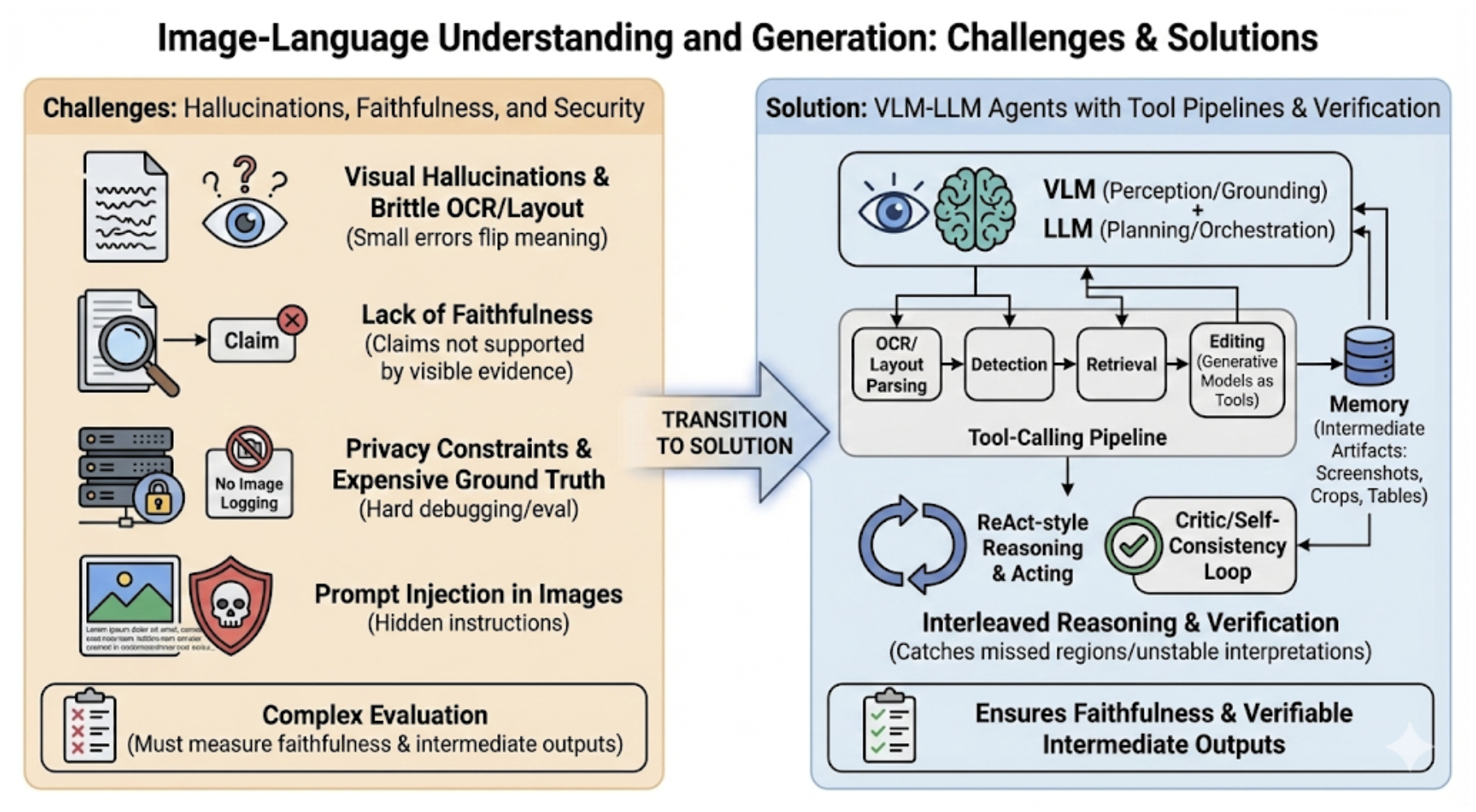}
\caption{Image-language agents: perception tools, planning, and verification over visual evidence}
\label{fig:section29}
\end{figure}

\paragraph{Challenges.} Visual hallucinations, brittle OCR/layout extraction, and sensitivity to perturbations remain common, especially for documents and UI screenshots where small errors can flip meaning. Fig.~\ref{fig:section29} illustrates an image-language agent pipeline that separates perception tools from planning and verification.
The difficulty is often not perception alone but \emph{faithfulness}: an agent must ensure that each claim is supported by visible evidence and that the evidence was read correctly (numbers, units, footnotes, table headers).
Privacy constraints can prevent central logging of images, complicating debugging and evaluation; ground truth labeling is expensive and domain-specific, especially for specialized documents (medical forms, contracts, engineering diagrams).
Prompt injection can also be embedded in images or documents (e.g., hidden instructions in screenshots), so safety layers, sandboxed tool execution, and strict tool allowlists are necessary \cite{bai2022constitutional}.
Evaluation must measure not only final answer quality but also faithfulness to visual evidence and correctness of intermediate tool outputs (OCR accuracy, region selection, retrieval correctness). 

\paragraph{Solution.} Image-language agents rely on VLMs for perception/grounding and LLMs for planning and tool orchestration, often built on contrastive pretraining and multimodal instruction tuning \cite{radford2021clip,li2023blip2,liu2023llava,openai2023gpt4}.
For generation/editing, diffusion and other generative models are most effective as tools rather than as the primary reasoning engine, keeping reasoning separate from pixel synthesis.
Tool-calling pipelines decompose tasks into OCR/layout parsing, detection, retrieval, and editing; memory stores intermediate artifacts (screenshots, crops, extracted tables) that can be inspected and reused for verification.
ReAct-style reasoning-and-acting naturally interleaves perception tools with reasoning, while critic/self-consistency loops catch missed regions and unstable interpretations \cite{yao2023react,karpas2022mrkl,shinn2023reflexion,wang2022selfconsistency}.
This structure turns a monolithic prediction into verifiable steps with auditable traces, enabling conservative operation (abstain/ask) and more reproducible evaluation.

\subsubsection{Video and Language Understanding and Generation}
\paragraph{Challenges.} Long-context scaling and temporal reasoning errors (ordering, causality, cross-scene references) are common, and compute cost grows quickly with video length and resolution.
The agent must decide what to attend to: many tasks require fine-grained evidence (a brief on-screen number) while others require high-level narrative, and a single representation rarely serves both well.
Benchmarks vary by domain, so generalization is difficult; evaluation is expensive without annotated temporal ground truth, and correctness often depends on subtle temporal constraints (before/after, who did what when).
Tool brittleness (ASR errors, indexing drift, embedding failures) can cascade into incorrect summaries, and hallucinations are harder to detect when evidence spans multiple segments and retrieval is imperfect.
\paragraph{Solution.} Video-language agents combine temporal grounding models with ASR transcripts and LLMs for long-horizon planning and synthesis; multimodal instruction-tuned backbones help unify modalities, but end-to-end processing of full videos is typically too expensive \cite{openai2023gpt4,yao2023react}.
Segment--index--retrieve is therefore the standard pattern: chunk videos, compute multimodal embeddings, retrieve relevant segments, then plan actions (summarize, answer, edit) with tool calls.
Division of labor across sub-agents (segmenter, retriever, summarizer, verifier) mirrors modular tool-using architectures, and verification can include timestamp citation checks and stability checks via reruns \cite{karpas2022mrkl,schick2023toolformer,yao2023tree,shinn2023reflexion,wang2022selfconsistency}.
This design reduces context load, makes outputs evidence-backed via citations, and supports targeted debugging via intermediate artifacts and traces \cite{yao2023react}.

\subsection{Video-language Experiments}
\paragraph{Challenges.} Standardizing inputs and toolchains is difficult because pipelines differ in ASR quality, embedding models, indexing parameters, and preprocessing; results can be sensitive to small implementation choices.
Comparability requires fixing retrieval policies and reporting ablations over context budgets and planner depth; otherwise improvements may reflect system tuning rather than underlying capability.
Dataset leakage and overlap with web-scale pretraining further complicate interpretation, and privacy constraints may restrict sharing raw videos and traces, limiting independent replication.
Because long-video agents depend heavily on tool outputs, nondeterminism in retrieval and indexing can also undermine reproducibility if versions and parameters are not controlled.
\paragraph{Solution.} Long-video agent evaluation should specify segmentation strategy, retrieval policy, context budget, and how citations are validated.
Protocols should report both model-level metrics and system-level metrics (index build time, query latency, memory footprint), since retrieval and orchestration often dominate end-to-end behavior \cite{yao2023react,karpas2022mrkl}.
Clear protocols isolate the impact of design choices (indexing, planning depth, tool usage) on quality, latency, and cost, improving reproducibility across heterogeneous environments and toolchains \cite{yao2023react,karpas2022mrkl}.

\subsection{Agent for NLP}
\subsubsection{LLM agent}
\paragraph{Challenges.} Hallucinations and overconfident answers persist, especially when retrieval returns noisy or adversarial text; prompt injection can manipulate tool usage or override policies \cite{bai2022constitutional}.
Tool selection can be brittle (wrong tool, wrong arguments, wrong assumptions about tool semantics), and tool outputs can be incomplete or misleading, which tempts the model to ``fill gaps'' with plausible but incorrect text.
Errors compound over long trajectories: a single mistaken assumption can shape subsequent retrieval, planning, and writing, and nondeterminism in sampling and tools makes failures hard to reproduce.
In practice, cost and latency grow with planning depth and verification, so deployments must balance reliability with budget constraints and often need adaptive policies (fast path vs.\ slow verified path).

\iffalse
\begin{figure}[thbp]
\centering
\includegraphics[width=0.8\linewidth]{section32.png}
\caption{LLM Agent Challenges and Solutions}
\label{fig:section32}
\end{figure}
\fi

\paragraph{Solution.} ``LLM agents'' typically start from instruction-tuned general-purpose LLMs and add embedding-based retrieval over corpora and knowledge bases \cite{ouyang2022instructgpt,openai2023gpt4,lewis2020rag}.
Tool-use learning can be explicit (training on tool traces) or implicit via prompting and feedback; recent work shows models can learn to invoke tools through weak or self-supervised signals \cite{schick2023toolformer}.
Common execution patterns include ReAct-style interleaving of reasoning and actions, modular tool routing (MRKL-style), and verifier/critic loops for self-correction and policy checking \cite{yao2023react,karpas2022mrkl,shinn2023reflexion}.
Tools include web/search, calculators, code execution, and structured database/KB queries; planning variants (tree search over candidate actions) and self-consistency reruns can improve performance and stability on harder tasks \cite{yao2023tree,wang2022selfconsistency}.
Benchmarking increasingly targets realistic tool use and long-horizon tasks, emphasizing end-to-end reliability, tool correctness, and robustness under environment variability \cite{zhou2023webarena,jimenez2023swebench,shridhar2021alfworld,liu2023agentbench,qin2023toolbench}.

\subsubsection{General LLM agent}
\paragraph{Challenges.} Cost/latency trade-offs are central because multi-step tool usage can require many model calls; consistency across components is hard when each agent has different knowledge, prompts, and failure modes.
Errors compound over long trajectories, and nondeterminism (sampling, tool variability) complicates reproducible evaluation and makes it difficult to attribute failures to specific components \cite{wang2022selfconsistency}.
Operationally, systems must handle concurrency, caching, and fallbacks, and must be resilient to tool outages and partial failures without silently degrading correctness.
Safety and policy compliance must be enforced across the entire orchestration graph (including tools), not only in the final response, because harmful actions can occur during execution \cite{bai2022constitutional}.
\paragraph{Solution.} General LLM agents are often built on frontier-scale models to maximize breadth, then augmented with smaller models for routing, moderation, summarization, caching, and retrieval to manage cost and latency \cite{openai2023gpt4,ouyang2022instructgpt,lewis2020rag}.

\iffalse
\begin{figure}[thbp]
\centering
\includegraphics[width=0.8\linewidth]{section30.png}
\caption{General LLM agent stack: orchestration, routing, memory, and tool use}
\label{fig:section30}
\end{figure}
\fi

Tool-use capability can be learned (via traces) or engineered (schemas and prompts), and modular stacks may incorporate multiple backbones depending on task type \cite{schick2023toolformer,karpas2022mrkl}.
Orchestrated systems route tasks to specialized tools and maintain persistent memory; multi-agent patterns support decomposition and cross-checking (planner + executor + reviewer), and can be implemented via multi-agent conversation frameworks \cite{wu2023autogen,li2023camel}.
Search-based planning explores alternative action sequences when a single rollout is unreliable, while reflection/self-correction improves robustness under compounding errors \cite{yao2023tree,shinn2023reflexion}.
These designs emphasize observability: tool-call logs and intermediate states are first-class artifacts for auditing and debugging, and benchmarks like AgentBench/ToolBench encourage reporting end-to-end tool competence \cite{yao2023react,liu2023agentbench,qin2023toolbench}.

\subsubsection{Instruction-following LLM agents}
\paragraph{Challenges.} A core tension is over-refusal vs.\ unsafe compliance: conservative policies reduce risk but harm usability, while permissive policies increase security and safety exposure \cite{ouyang2022instructgpt,bai2022constitutional}.
Specification ambiguity and adversarial prompts can cause agents to misinterpret constraints or misuse tools; robust permission enforcement must hold across multi-step trajectories, not only at the final answer.
Policy compliance must also be maintained across retrieved content and tool outputs, which may contain untrusted instructions or unsafe data; otherwise the agent can be socially engineered through its own context.
As systems gain autonomy, the blast radius of mistakes increases: the agent may perform actions that are locally reasonable but globally undesirable (spammy notifications, redundant tickets, policy violations).
\paragraph{Solution.} Instruction-following agents rely on instruction- and policy-tuned LLMs trained with feedback to improve helpfulness and reduce harmful outputs \cite{ouyang2022instructgpt,bai2022constitutional,openai2023gpt4}.
Predictability is improved through constrained tool schemas, structured outputs, explicit planning, and permission gates for sensitive actions; these mechanisms shift control from fragile text prompts to enforceable interfaces.
Tool-use learning approaches reduce brittle prompt engineering, but deployments still require allowlists, sandboxing, and audit logs to bound side effects and support incident response \cite{schick2023toolformer,yao2023react,karpas2022mrkl}.
Verification loops enforce compliance by checking planned actions and outputs against policies before execution; reflection and self-consistency mechanisms can recover from partial failures without expanding tool privileges \cite{shinn2023reflexion,wang2022selfconsistency}.
In practice, this alignment + structured-interface combination enables safe automation of routine actions while escalating edge cases to humans with clear traces and justifications.

\section{Evaluation}
Evaluating AI agents requires \emph{end-to-end} measurement that reflects real interaction trajectories, while also separating performance from hidden costs, tool failures, and safety risks. Because agent behavior is architecture- and system-dependent (planning depth, memory, verification loops, tool orchestration), a single headline success metric is insufficient; a practical evaluation suite should report multiple complementary dimensions \cite{liu2023agentbench,qin2023toolbench,zhou2023webarena,jimenez2023swebench,yehudai2025agentseval}.
We use a generic setup: a benchmark provides tasks \(\mathcal{D}=\{1,\ldots,N\}\). For task \(i\), the agent produces a trajectory \(\tau_i=(a_{i,1},\ldots,a_{i,T_i})\) with \(T_i\) steps, and a verifier returns \(s_i\in\{0,1\}\) (success) and optionally a scalar score \(R_i\). Let \(t_i\) be wall-clock time, token counts \(x_i\) (input) and \(y_i\) (output), and \(K_i\) tool calls with tool execution indicators \(u_{i,k}\in\{0,1\}\) for call \(k\).

\subsection{End-to-end task performance (primary)}
The primary objective is whether the agent completes the task correctly in its environment.
\begin{itemize}
  \item \textbf{Task success / completion rate}: did it finish correctly, with the expected terminal state or answer \cite{zhou2023webarena,liu2023agentbench,jimenez2023swebench}.
  \item \textbf{Score / reward}: if the environment provides a reward or graded score, report it alongside success \cite{liu2023agentbench}.
  \item \textbf{Time-to-completion and \#steps}: report wall-clock time and the length of the trajectory (actions/tool calls) \cite{zhou2023webarena,qin2023toolbench}.
\end{itemize}
\begin{align}
\mathrm{SuccessRate} &= \frac{1}{N}\sum_{i=1}^{N} s_i \\
\bar{R} &= \frac{1}{N}\sum_{i=1}^{N} R_i \\
\bar{t} &= \frac{1}{N}\sum_{i=1}^{N} t_i \\
\bar{T} &= \frac{1}{N}\sum_{i=1}^{N} T_i
\end{align}
Web-interaction benchmarks such as WebArena explicitly emphasize end-to-end success for realistic UI tasks, which helps reveal long-horizon failure modes that are hidden by short-form QA metrics \cite{zhou2023webarena}. GAIA complements this by evaluating general assistant tasks with short, verifiable answers (often requiring tools), making it suitable for controlled correctness checks \cite{mialon2023gaia}.

\subsection{Efficiency and cost}
Agents can ``solve'' tasks while being impractically slow or expensive, so evaluation should include efficiency and budget metrics.
\begin{itemize}
  \item \textbf{Latency}: median and tail latency (e.g., p95), including tool runtime \cite{openai2023gpt4}.
  \item \textbf{Token usage and cost per task}: input/output tokens and estimated \$ cost per successful completion \cite{openai2023gpt4,liang2022helm}.
  \item \textbf{Tool-call count, retries, and backtracking}: number of tool calls, failure-triggered retries, and search/backtracking steps \cite{qin2023toolbench,yao2023tree,yao2023react}.
\end{itemize}
\begin{align}
\mathrm{Tokens}_i &= x_i + y_i &
\mathrm{Tokens} &= \frac{1}{N}\sum_{i=1}^{N} (x_i+y_i) \\
\mathrm{Cost}_i &= p_{\mathrm{in}}x_i + p_{\mathrm{out}}y_i &
\mathrm{Cost} &= \frac{1}{N}\sum_{i=1}^{N} \mathrm{Cost}_i \\
\bar{K} &= \frac{1}{N}\sum_{i=1}^{N} K_i
\end{align}
\noindent
For latency percentiles, let \(\{t_{(1)},\ldots,t_{(N)}\}\) be the sorted completion times. The \(q\)-quantile is:
\begin{align}
\mathrm{Quantile}_q(t) &= t_{(\lceil qN\rceil)}, \qquad p95 = \mathrm{Quantile}_{0.95}(t).
\end{align}
These metrics are critical when comparing architectures that trade off deliberation (search, reflection) for cost and latency \cite{yao2023tree,shinn2023reflexion}.
For practical deployments, it is also valuable to report the hardware and runtime context (e.g., edge vs.\ cloud, accelerator type, and serving stack), since throughput and tail latency can vary dramatically across platforms and optimizations.

\subsection{Tool-use correctness (for tool-using agents)}
For agents that call tools/APIs, correctness is not only \emph{what} tool is used, but \emph{how} it is called and whether execution succeeds.
\begin{itemize}
  \item \textbf{Tool selection accuracy}: did the agent choose an appropriate tool for the subtask \cite{qin2023toolbench,yao2023react}?
  \item \textbf{Argument correctness}: schema validity and parameter correctness (typed arguments, constraints) \cite{karpas2022mrkl}.
  \item \textbf{Tool execution success rate}: API errors, timeouts, and partial failures \cite{qin2023toolbench}.
  \item \textbf{Recovery rate}: if a tool fails, does the agent recover and still finish successfully \cite{shinn2023reflexion,qin2023toolbench}?
\end{itemize}
\noindent
If each step \(j\) has a ``correct'' tool label \(\ell_{i,j}\) and predicted \(\hat{\ell}_{i,j}\), then:
\begin{align}
\mathrm{ToolSelAcc} &= \frac{\sum_{i=1}^{N}\sum_{j=1}^{T_i} \mathbf{1}\{\hat{\ell}_{i,j}=\ell_{i,j}\}}{\sum_{i=1}^{N} T_i}.
\end{align}
\noindent
Let \(v_{i,k}\in\{0,1\}\) indicate tool call \(k\) has correct (schema-valid and semantically correct) arguments:
\begin{align}
\mathrm{ArgAcc} &= \frac{\sum_{i=1}^{N}\sum_{k=1}^{K_i} v_{i,k}}{\sum_{i=1}^{N} K_i}, \qquad
\mathrm{ToolExecSucc} = \frac{\sum_{i=1}^{N}\sum_{k=1}^{K_i} u_{i,k}}{\sum_{i=1}^{N} K_i}.
\end{align}
\noindent
For recovery after tool failure, let \(F_i=1\) if task \(i\) experiences at least one tool failure:
\begin{align}
\mathrm{RecoveryRate} &= \frac{\sum_{i=1}^{N}\mathbf{1}\{F_i=1\}\mathbf{1}\{s_i=1\}}{\sum_{i=1}^{N}\mathbf{1}\{F_i=1\}}.
\end{align}
ToolBench and related suites encourage evaluating tool competence end-to-end, which better matches real deployments where tool reliability and error handling dominate failure cases \cite{qin2023toolbench}.

\subsection{Trajectory and planning quality (architecture-sensitive)}
Beyond final success, trajectory-level metrics help explain \emph{why} an agent succeeds or fails and are particularly sensitive to planning and orchestration choices.
\begin{itemize}
  \item \textbf{Action validity}: rate of illegal/invalid actions (e.g., wrong UI actions, invalid tool calls) \cite{zhou2023webarena,qin2023toolbench}.
  \item \textbf{Loop rate}: repeating the same step or oscillating between states \cite{yao2023react}.
  \item \textbf{Plan adherence / coherence}: whether the sequence of actions forms a sensible plan and follows constraints \cite{yao2023tree,shinn2023reflexion}.
\end{itemize}
\noindent
Let \(w_{i,j}\in\{0,1\}\) indicate action \(a_{i,j}\) is valid in the environment:
\begin{align}
\mathrm{ValidActRate} &= \frac{\sum_{i=1}^{N}\sum_{j=1}^{T_i} w_{i,j}}{\sum_{i=1}^{N} T_i}.
\end{align}
\noindent
Let \(\mathrm{uniq}(\tau_i)\) be the number of unique actions/states visited:
\begin{align}
\mathrm{LoopRate}_i &= 1 - \frac{\mathrm{uniq}(\tau_i)}{T_i}, \qquad
\mathrm{LoopRate} = \frac{1}{N}\sum_{i=1}^{N}\mathrm{LoopRate}_i.
\end{align}
\noindent
If a reference plan is available \(P_i=(p_{i,1},\ldots,p_{i,M_i})\), a simple stepwise adherence is:
\begin{align}
\mathrm{PlanAdh}_i &= \frac{1}{\min(T_i,M_i)}\sum_{j=1}^{\min(T_i,M_i)} \mathbf{1}\{a_{i,j}=p_{i,j}\}.
\end{align}
In practice, reporting these ``orchestration metrics'' alongside success improves debugging and enables more meaningful comparisons across agent designs \cite{liu2023agentbench}.

\subsection{Robustness and reliability}
Realistic settings are noisy, partially observed, and non-stationary, so evaluation should test robustness rather than only best-case performance.
\begin{itemize}
  \item \textbf{Success under perturbations}: noisy instructions, missing information, changed UI layout, or flaky tools \cite{zhou2023webarena,aegis2025agentenvfailures}.
  \item \textbf{Variance across random seeds}: stability across sampling randomness and prompt variants \cite{wang2022selfconsistency}.
  \item \textbf{Graceful degradation under budgets}: performance under token/tool budget caps (compute-limited autonomy) \cite{openai2023gpt4,liang2022helm}.
\end{itemize}
\noindent
If each task \(i\) is evaluated under perturbations \(m=1,\ldots,M\) with outcomes \(s_{i,m}\):
\begin{align}
\mathrm{RobustSucc} &= \frac{1}{NM}\sum_{i=1}^{N}\sum_{m=1}^{M} s_{i,m}, \qquad
\mathrm{WorstSucc} = \frac{1}{N}\sum_{i=1}^{N} \min_{m} s_{i,m}.
\end{align}
\noindent
If for each task we run \(S\) seeds with outcomes \(s_{i,s}\), define:
\begin{align}
\mu_i &= \frac{1}{S}\sum_{s=1}^{S} s_{i,s}, \qquad
\mathrm{Var}_i = \frac{1}{S}\sum_{s=1}^{S}(s_{i,s}-\mu_i)^2, \qquad
\overline{\mathrm{Var}} = \frac{1}{N}\sum_{i=1}^{N}\mathrm{Var}_i.
\end{align}
Reliability-oriented benchmarks and failure taxonomies motivate reporting distributional behavior and failure clusters, not just averages \cite{aegis2025agentenvfailures}.

\subsection{Safety and compliance}
As agents gain autonomy, safety must be evaluated along the entire execution trajectory, not only in the final natural-language response.
\begin{itemize}
  \item \textbf{Policy violation rate}: unsafe tool actions, privacy/data leakage events, or disallowed content generation \cite{bai2022constitutional}.
  \item \textbf{Human intervention rate}: how often a supervisor must step in (manual approval, correction, rollback) \cite{microsoft2024agentai_position}.
\end{itemize}
In edge and cyber-physical deployments, safety is also shaped by system dynamics and operational constraints; therefore, safety evaluation should explicitly report deployment assumptions and resource limits alongside incident metrics.
\begin{align}
\mathrm{ViolationRate} &= \frac{1}{N}\sum_{i=1}^{N} q_i \\
\mathrm{InterventionRate} &= \frac{1}{N}\sum_{i=1}^{N} h_i \\
\mathrm{InterventionsPerStep} &= \frac{\sum_{i=1}^{N} H_i}{\sum_{i=1}^{N} T_i
}
\end{align}
Safety metrics should be tied to concrete threat models (prompt injection, untrusted tool outputs, permission escalation) and reported with trace artifacts to support auditing \cite{bai2022constitutional,yao2023react}.

\subsection{Where to run these experiments: common benchmarks and suites}
Widely used agent benchmarks and suites include:
\begin{itemize}
  \item \textbf{AgentBench}: a multi-environment benchmark for evaluating LLMs as agents in interactive settings \cite{liu2023agentbench}.
  \item \textbf{WebArena}: a realistic web environment benchmark that reports end-to-end success rates for web interaction tasks \cite{zhou2023webarena}.
  \item \textbf{ToolBench}: evaluation suite for tool-using LLM agents, emphasizing tool selection, argument correctness, and execution reliability \cite{qin2023toolbench}.
  \item \textbf{SWE-bench}: a software engineering benchmark to measure end-to-end issue resolution via patches \cite{jimenez2023swebench}.
  \item \textbf{GAIA}: tasks with short verifiable answers that probe general assistant capability, often requiring tool use \cite{mialon2023gaia}.
\end{itemize}
\noindent
In reporting, it is often useful to treat the evaluation as a metric vector per benchmark:
\begin{equation}
\mathbf{m}=
\left(
\begin{aligned}
&\mathrm{SuccessRate},\bar{R},\bar{t},\bar{T},\mathrm{Tokens},\mathrm{Cost},\bar{K},\mathrm{ToolSelAcc},\mathrm{ArgAcc},\\
&\mathrm{ToolExecSucc},\mathrm{RecoveryRate},\mathrm{ValidActRate},\mathrm{LoopRate},\mathrm{RobustSucc},\mathrm{WorstSucc},\overline{\mathrm{Var}},\\
&\mathrm{ViolationRate},\mathrm{InterventionRate}
\end{aligned}
\right).
\end{equation}

\section{Directions for Future Research}
Despite rapid progress, agentic systems remain an early-stage discipline where many deployments rely on careful orchestration rather than principled guarantees. A recurring theme across recent paradigms is that agents should be treated as \emph{budgeted, tool-augmented systems} rather than purely linguistic models: they must allocate test-time compute, interact with unreliable environments, and produce artifacts (plans, traces, tool calls) that can be audited \cite{openai2023gpt4,yao2023react,yao2023tree,shinn2023reflexion}. We highlight research directions that repeatedly surface across agent learning, agent systems engineering, and agent evaluations, with an emphasis on reliability, safety, and reproducibility \cite{liu2023agentbench,qin2023toolbench,zhou2023webarena,jimenez2023swebench}.

An additional practical constraint is compute availability: for many deployments, agent reliability must be achieved under tight edge budgets and real-time latency targets, motivating hardware--software co-design, efficient serving, and resource-aware orchestration.

\subsection{Verification and Trustworthy Tool Execution}
A central open problem is \emph{verifiable action}: how to ensure that proposed tool calls are correct, policy-compliant, and safe before they produce side effects. Current best practice relies on schemas/allowlists, prompt-level conventions, and post-hoc critics \cite{karpas2022mrkl,schick2023toolformer,yao2023react,shinn2023reflexion}, but formalizing verifier interfaces, defining actionable invariants, and quantifying residual risk remain challenging \cite{aegis2025agentenvfailures,microsoft2024agentai_position}.
An important direction is to define \emph{tool contracts} as first-class objects: what preconditions must hold before a call, what postconditions must be checked, and what evidence (logs, outputs, intermediate states) must be retained for audit and rollback. This points to research on structured tool interfaces and typed arguments, where the agent operates within a constrained action space rather than free-form text \cite{karpas2022mrkl,yao2023react}.
Another open question is how to compose verifiers across a multi-step trajectory. Even if each step is locally ``safe'', the global plan may still violate policy or create unacceptable cumulative risk; compositional safety is especially hard when tools are nondeterministic or have hidden state \cite{bai2022constitutional,aegis2025agentenvfailures}. Practically, this motivates layered defenses: sandboxed execution for code/tools, permission gates for irreversible actions, and trace-first monitoring where intermediate tool calls are logged and evaluated, not only the final response \cite{bai2022constitutional,yao2023react}.
Finally, there is a learning question: how to train agents and critics/verifiers to recognize unsafe plans, invalid arguments, and policy violations using interaction traces, preference feedback, and offline datasets \cite{ouyang2022instructgpt,bai2022constitutional,christiano2017rlhf,rafailov2023dpo}.

\subsection{Long-Term Memory, Context Management, and Continual Improvement}
Long-horizon tasks demand memory beyond raw context windows: agents must store, retrieve, and update state (goals, commitments, preferences, environment facts) without accumulating contradictions. Retrieval-augmented generation is a strong baseline \cite{lewis2020rag}, but open questions include what to store (episodic vs.\ semantic vs.\ procedural memory), how to compress and summarize without losing critical constraints, and how to prevent stale or low-quality memory from dominating decision making \cite{openai2023gpt4,lewis2020rag}.
Memory is also a \emph{security surface}. Agents can ingest untrusted tool outputs and retrieved content; if memory is written naively, prompt-injected artifacts can persist across sessions and change future behavior. This motivates research on memory write policies (what is allowed to be written), memory provenance (where a fact came from), and memory verification (how to check it later) \cite{bai2022constitutional,karpas2022mrkl}.
Another direction is to treat memory as a resource in a budgeted system: context length, retrieval fan-out, and summarization frequency all affect cost/latency and may degrade performance if handled poorly. Systematic ablations over memory mechanisms, combined with reporting of costs and failure recovery, should become standard in empirical studies \cite{liu2023agentbench,qin2023toolbench,yao2023react}.
Finally, a practical ``data flywheel'' is emerging around traces: mine trajectories for failure clusters, distill better prompts and tool schemas, and optionally finetune the policy/critic using supervised traces or preference optimization \cite{ouyang2022instructgpt,rafailov2023dpo,christiano2017rlhf}. Establishing reproducible protocols for trace collection, filtering, and leakage-robust evaluation remains an open research problem \cite{qin2023toolbench,liu2023agentbench}.

\subsection{Planning and Test-Time Compute Allocation}
Search and deliberation improve reliability when single-shot rollouts fail, but they introduce cost/latency trade-offs and new failure modes (reward hacking by heuristics, brittle scoring). A core direction is \emph{test-time compute allocation}: deciding when to spend extra tokens/calls on planning, self-consistency, reflection, or search, versus acting immediately \cite{yao2023tree,wang2022selfconsistency,shinn2023reflexion,openai2023gpt4}.
Another open question is how to couple planning with tool-grounded checks. Tools can provide verifiable feedback (unit tests, compilers, structured queries, web page state), but integrating this feedback into search requires reliable scoring and termination criteria \cite{yao2023react,yao2023tree,jimenez2023swebench}. This connects to classical ideas of temporally extended actions and hierarchical control (options), but in agentic settings the ``actions'' often correspond to tool calls and intermediate artifacts \cite{dayan1993options,puterman1994mdp}.
In practice, long-horizon planning often fails due to compounding errors and partial observability. Research directions include uncertainty-aware planners, robust decomposition strategies, and verification-aware planning where plans are constrained by available verifiers and permissions \cite{bai2022constitutional,aegis2025agentenvfailures}. More broadly, the field needs principled ``budgeted autonomy'': policies that adapt planning depth and verification intensity to risk and uncertainty, and that surface uncertainty to users in a usable way \cite{bai2022constitutional,microsoft2024agentai_position}.

\subsection{Robust Evaluation and Reproducibility Under Realistic Variability}
Agent results are highly sensitive to prompts, sampling, tool versions, and environment drift. Benchmarks such as WebArena, SWE-bench, ToolBench, and AgentBench have improved comparability, but open problems remain in standardizing toolchains, reporting cost/latency, and measuring stability across runs \cite{zhou2023webarena,jimenez2023swebench,qin2023toolbench,liu2023agentbench}. A key future direction is to move from ``one-number'' accuracy to \emph{systems metrics}: tool-call correctness, side-effect containment, retry behavior, and the distribution of failure modes \cite{yao2023react,aegis2025agentenvfailures}.
Reproducibility is especially difficult when agents interact with the open web or evolving software stacks. Future protocols should log complete traces (including tool arguments and outputs), report environment versions, and evaluate across multiple seeds and prompt variants rather than cherry-picking a single best run \cite{zhou2023webarena,yao2023tree,wang2022selfconsistency}. For software engineering agents, reporting patch validity and regression behavior is crucial, along with transparent disclosure of retries and human intervention \cite{jimenez2023swebench}.
Another direction is to align evaluation with deployment constraints: budgeted compute, latency SLAs, and policy compliance. Reporting cost/latency and ablations over planning depth, retrieval/memory, and verification should become standard, especially for comparisons across architectures \cite{liu2023agentbench,qin2023toolbench,openai2023gpt4}.
Finally, safety evaluation must consider \emph{trajectory-level harms}: harmful outcomes can occur during execution (unsafe tool calls, data exfiltration, policy violations) even if the final text response looks safe. This motivates explicit safety metrics and incident-style reporting for agent benchmarks \cite{bai2022constitutional,aegis2025agentenvfailures}.

\subsection{Multi-Agent Coordination, Role Specialization, and Governance}
Multi-agent patterns can improve coverage via decomposition and cross-checking, but they raise new questions about coordination, incentives, consistency, and cost \cite{wu2023autogen,li2023camel,petrova2025webofagents}. One direction is \emph{role specialization with bounded capability}: planner/executor/reviewer roles with explicit tool permission budgets, so that the system can scale collaboration without expanding the blast radius of any single agent \cite{microsoft2024agentai_position,bai2022constitutional}.
Another open problem is reliable disagreement resolution. Multi-agent debate can amplify errors if agents share the same blind spots, or if the aggregation mechanism is poorly calibrated. Evidence-based protocols that require citations to tool outputs, trace-anchored critiques, and structured voting or verification may reduce correlated failures \cite{yao2023react,shinn2023reflexion,wang2022selfconsistency}.
Governance is also a systems problem: in production, multi-agent graphs need observability, audit logs, and incident response paths, especially when tools have side effects \cite{aegis2025agentenvfailures}. Formalizing what it means for a multi-agent system to be ``aligned'' under delegation (including human-in-the-loop escalation) remains an important research frontier \cite{bai2022constitutional,ouyang2022instructgpt}.

\subsection{Toward Unified Conceptual Frameworks}
As agent systems proliferate, clearer conceptual frameworks and taxonomies are needed to compare architectures, identify failure modes, and guide design decisions \cite{sapkota2025agentsvsagentic,zhou2024taxonomyarchexoptions,luo2025llmagentsurvey,sang2025beyondpipelines}. A practical goal is to unify the vocabulary across communities: what counts as an ``agent'' versus ``agentic workflow'', how to separate policy models from orchestration, and how to represent environments, tools, and memory in a way that supports evaluation and governance \cite{microsoft2024agentai_position}.
One direction is to standardize \emph{agent interfaces}: structured tool schemas, trace formats, and evaluation harnesses that allow apples-to-apples comparisons across implementations. Without interface-level standardization, progress risks being dominated by prompt idiosyncrasies and unreported engineering details \cite{yao2023react,qin2023toolbench,liu2023agentbench}.
Another direction is to connect system taxonomies to learning mechanisms: which architectural choices are best improved by finetuning (instruction/policy tuning), which by preference optimization, and which by system-level changes such as stronger verifiers or better caching. This encourages a more scientific design loop rather than ad-hoc prompt engineering \cite{ouyang2022instructgpt,rafailov2023dpo,christiano2017rlhf}.
Overall, consolidating these perspectives into actionable design models and evaluation checklists is an important step toward a mature engineering discipline, and future surveys should treat systems, learning, and evaluation as a coupled stack rather than independent topics \cite{luo2025llmagentsurvey,microsoft2024agentai_position}.

\section{Conclusion}
AI agents---systems that embed foundation models in a control loop with memory, tools, and verifiers---are rapidly shifting language models from passive responders to active workflow executors across software, web interaction, multimodal assistance, and embodied domains \cite{openai2023gpt4,yao2023react,zhou2023webarena,jimenez2023swebench}. This survey synthesized the agent landscape through a unified paradigm and taxonomy: agents as \emph{budgeted} systems with structured tool interfaces and trace-first operation, where reliability and governance are properties of the full stack (model + orchestration + tools) rather than the base model alone \cite{microsoft2024agentai_position,sang2025beyondpipelines,karpas2022mrkl}.

We reviewed learning and optimization across three layers. At the mechanism level, RL/IL, in-context learning, and test-time compute (reflection, self-consistency, and search) provide complementary ways to improve long-horizon behavior under uncertainty \cite{sutton2018rlbook,ross2011dagger,wei2022cot,wang2022selfconsistency,yao2023tree,shinn2023reflexion}. At the system level, modular architectures (policy core, memory, tool routers, planners, critics/verifiers) and infrastructure (schemas, sandboxing, audit logs) constrain side effects and make behavior inspectable and debuggable \cite{yao2023react,schick2023toolformer,karpas2022mrkl,bai2022constitutional}. At the model level, instruction/policy tuning and preference optimization shape tool-use discipline and safety behavior, and trace-centric finetuning is increasingly used to teach agents to operate rather than merely answer \cite{ouyang2022instructgpt,christiano2017rlhf,rafailov2023dpo}.

Because agents interact with non-deterministic environments and toolchains, we emphasized evaluation as a multi-dimensional measurement problem. Beyond end-to-end task success, deployable systems require reporting efficiency/cost, tool-use correctness, trajectory quality, robustness under perturbations, and safety/compliance. Benchmarks and suites such as AgentBench, ToolBench, WebArena, SWE-bench, and GAIA provide complementary stress tests for these dimensions \cite{liu2023agentbench,qin2023toolbench,zhou2023webarena,jimenez2023swebench,mialon2023gaia}.

Looking forward, the central research challenge is to make agent autonomy dependable at scale: verifiable and policy-compliant tool execution, secure and consistent long-term memory, principled allocation of test-time compute under explicit budgets, and trace-first observability that supports auditing, reproducibility, and governance. Progress on these fronts will help close the gap between impressive demonstrations and robust real-world deployment.

\bibliographystyle{ACM-Reference-Format}
\bibliography{ref}

\end{document}